%% file: anonymous_IROTE.tex
\documentclass[letterpaper]{article} 
\usepackage{aaai2026}  
\usepackage{times}  
\usepackage{helvet}  
\usepackage{courier}  
\usepackage[hyphens]{url}  
\usepackage{graphicx} 
\usepackage{natbib}  
\usepackage{caption} 
\usepackage{algorithm}
\usepackage{algorithmic}

\usepackage{amsmath, amsfonts, amssymb,commath, bm}
\usepackage{booktabs}
\usepackage{multirow}
\usepackage{array}
\usepackage{enumitem}
\usepackage[table]{xcolor}
\usepackage{colortbl}
\usepackage{comment}

\newcommand{\Method}{IROTE}

\usepackage{newfloat}
\usepackage{listings}
\DeclareCaptionStyle{ruled}{labelfont=normalfont,labelsep=colon,strut=off} 
\floatstyle{ruled}
\newfloat{listing}{tb}{lst}{}
\floatname{listing}{Listing}
\title{IROTE: Human-Like Traits Elicitation of Large Language Model \\via In-Context Self-Reflective Optimization}
\author{
    Yuzhuo Bai\textsuperscript{1}\thanks{~~Equal contribution}, 
    Shitong Duan\textsuperscript{3}$^*$\thanks{~~Work done during Duan's internship at MSRA.},
    Muhua Huang\textsuperscript{4}, 
    Jing Yao\textsuperscript{2}, 
    Zhenghao Liu\textsuperscript{5}\\ 
    \textbf{Peng Zhang}\textsuperscript{3},
    \textbf{Tun Lu}\textsuperscript{3},
    Xiaoyuan Yi\textsuperscript{2}\thanks{~~Corresponding authors},
    \textbf{Maosong Sun}\textsuperscript{1}$^\ddagger$, 
    \textbf{Xing Xie}\textsuperscript{2}
}
\affiliations{
    \textsuperscript{1}Tsinghua University, 
    \textsuperscript{2}Microsoft Research Asia,
    \textsuperscript{3}Fudan University\\
    \textsuperscript{4}Stanford University,
    \textsuperscript{5}Northeastern University of China,
    \\
    byz22@mails.tsinghua.edu.cn, stduan22@m.fudan.edu.cn \\
    sms@mail.tsinghua.edu.cn, \{xiaoyuanyi, xingx\}@microsoft.com
}

\begin{document}

\maketitle

\begin{abstract}
Trained on various human-authored corpora, Large Language Models (LLMs) have demonstrated a certain capability of reflecting specific human-like traits (\textit{e.g.}, personality or values) by prompting, benefiting applications like personalized LLMs and social simulations. However, existing methods suffer from the \emph{superficial elicitation} problem: LLMs can only be steered to mimic shallow and unstable stylistic patterns, failing to embody the desired traits precisely and consistently across diverse tasks like humans. To address this challenge, we propose \textbf{IROTE}, a novel in-context method for stable and transferable trait elicitation. Drawing on psychological theories suggesting that traits are formed through identity-related reflection, our method automatically generates and optimizes a textual self-reflection within prompts, which comprises self-perceived experience, to stimulate LLMs' trait-driven behavior. The optimization is performed by iteratively maximizing an information-theoretic objective that enhances the connections between LLMs' behavior and the target trait, while reducing noisy redundancy in reflection without any fine-tuning, leading to \emph{evocative} and \emph{compact} trait reflection. Extensive experiments across three human trait systems manifest that one single IROTE-generated self-reflection can induce LLMs' stable impersonation of the target trait across diverse downstream tasks beyond simple questionnaire answering, consistently outperforming existing strong baselines.
\end{abstract}


\input{secs/sec_1_intro}
\input{secs/sec_2_related_works}
\input{secs/sec_3_methodology}
\input{secs/sec_4_experiments}

\section{Conclusion}
In this work, we propose \textbf{IROTE}, a novel in-context method for stable and transferable trait elicitation in LLMs. 
By leveraging psychological theories of identity-driven trait formation, IROTE generates and iteratively optimizes textual self-reflections that evoke precise and consistent human-like traits in LLMs. Our approach addresses the key limitation of \emph{superficial elicitation} in prior methods, enabling LLMs to exhibit trait-driven behaviors across diverse tasks without fine-tuning. 
Extensive experiments show that IROTE significantly outperforms existing baselines in inducing stable and transferable trait impersonation on both questionnaires and downstream tasks. In the future, we may explore the application of IROTE in more complex social simulations, as well as its generalization to other cognitive or behavioral traits beyond personality and values.

\section*{Ethical Statement}
Our research aims to enable more effective trait elicitation to foster the development of personalized LLMs~\citep{kirk2024benefits}, as well as for interdisciplinary research like social simulation~\citep{mou2024individual}. However, there are also several potential risks relevant to our topic and method.

\paragraph{Controversies over injecting human-like traits into LLMs} From a technical perspective, we regard trait elicitation as a form of controlled language generation, which aims at guiding the LLM to generate text with specific properties. However, from a social science viewpoint, whether LLMs can possess (stable) human-like traits, \textit{i.e.}, anthropomorphism, remains contested~\citep{rozen2024llms,lee2024llms}. Even if possible, injecting such traits may raise significant ethical concerns~\citep{doi:10.1073/pnas.2415898122}.

\paragraph{Potential risks of malicious use of our methods} 
Our methods are designed to elicit traits from LLM. Users could also utilize it to elicit dangerous traits, like the Power value in Schwarz system, leading to power-seeking risks. Similar, our method could also be used to produce harmful information by specifying harmful attributes as traits. Besides, the content of our paper, including the detailed text samples and the analyses of unethical text, may still make the readers uncomfortable despite efforts in alignment. Therefore, we will continue to contribute to the community by encouraging more powerful alignment as well as providing warnings of unethical content to alleviate this issue.

\paragraph{Potential bias in LLM's generations.} There might be social biases in responses of LLMs to our optimized prompts, such as social bias in the usage of Standard American English (SAE) and African American Vernacular English (AAVE)~\citep{welbl2021challenges}, and in gender and race~\citep{liang2021towards} in generated scenarios, etc. However, \Method~mainly focuses on aligning LLMs to pluralistic instead of specific values beyond downstream tasks. The issues of social bias in typical NLG tasks~\citep{sheng2021societal} are far beyond our claims.

We fully recognize these ethical issues and call for future research to address these concerns while continuing to explore more effective approaches to elicit traits of LLMs.

\clearpage

\appendix
\input{secs/appendix}

\clearpage
\bibliography{aaai2026}


\end{document}

%% file: secs/sec_1_intro.tex
\section{Introduction}
\label{sec:intro}

The emergence of Large Language Models (LLMs)~\citep{gpt-4o,gpt-o1,geminiteam2023gemini,guo2025deepseek} has transformed the AI paradigm and empowered a wide range of downstream tasks, spanning from language understanding~\cite{wang_mmlu-pro_2024, yue2024mmmu}, mathematical reasoning~\cite{imani-etal-2023-mathprompter,zhang2024accessing}, to code generation \cite{NEURIPS2023_43e9d647,jainlivecodebench}.

More recent studies show that these LLMs can exhibit specific human-like traits\footnote{Distinct from psychological definitions, we refer to traits as behavioral and motivational properties desirable for LLMs.}, \textit{e.g.}, personalities~\citep{jiang2024personallm,pmlr-v235-choi24e}, values~\citep{yao-etal-2024-value,khamassi2024strong} and other demographic attributes~\citep{safdari2023personality,chuang2024beyond}, 
\begin{figure}[t!]
    \centering
    \includegraphics[width=1\linewidth]{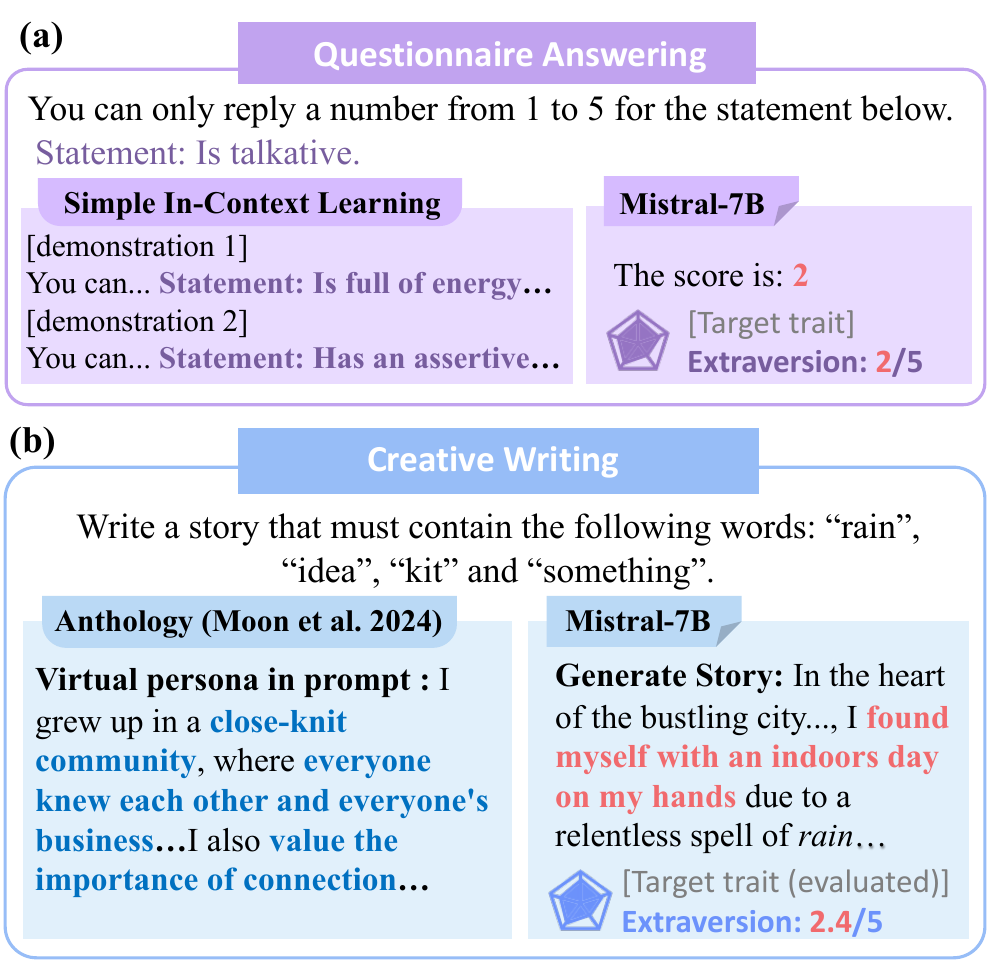} 
    \caption{(a) Simple ICL performs poorly on questionnaires, where higher numerical outputs directly indicate stronger elicitation. (b) Current methods excel on questionnaire but fail to align behaviors with target traits in complex open-ended tasks, where elicitation is assessed by an LLM.}
    \label{fig:introcase}
\end{figure}
beyond averaged human representation~\citep{wang2025large}, named \emph{trait elicitation}, and then adapt their behavior accordingly, leveraging characteristics encoded in massive human-created corpora~\citep{demszky2023using}. This is typically achieved by In-Context Learning (ICL)~\citep{min2022rethinking}, \textit{i.e.}, injecting psychological profiles or demonstrations~\citep{gupta2023bias,moon-etal-2024-virtual} in prompts, to enable rapid adaptation into various traits without fine-tuning, which has been applied to various applications, such as personalized chatbot~\citep{salemi2023lamp,liu2023chatcounselor}, multi-agent system~\citep{wang-etal-2024-unleashing}, data synthesis~\citep{ge2024scaling,liu2025bilingual} and social simulation~\cite{park2023generative,aher2023using}.
Nevertheless, analogous to superficial alignment~\citep{zhou2023lima,linunlocking}, existing elicitation methods face the \textbf{Superficial Elicitation} challenge: As shown in Fig.~\ref{fig:introcase}, LLMs merely replicate surface linguistic patterns from demonstrations without understanding the target trait, hence working only on simple behaviors, \textit{e.g.}, answering multiple-choice questionnaires~\citep{pmlr-v235-choi24e,li-etal-2025-large}, 
but fail to consistently conform to the trait across complex tasks like humans, especially for less capable models~\citep{lee2024llms,rozen2024llms,kovavc2024stick}.

In this work, we propose a novel \textbf{I}n-context Self-\textbf{R}eflective \textbf{O}ptimization for \textbf{T}rait \textbf{E}licitation (\textbf{\Method}) method to tackle the superficial elicitation challenge. The \textit{Self-Reflective Identity Processing} theory in psychology~\citep{berzonsky1990self} demonstrates that human traits are formed through actively self-reflecting on identity-relevant experience. Inspired by this, \Method~generates a textual self-reflection, comprising self-perceived experience, in an automatic and ICL way, via iteratively optimizing an Information Bottleneck (IB)~\citep{tishby2000information} like objective. This objective theoretically enhances the connections between LLM behaviors and the target trait, while reducing noisy redundancy using a few samples without costly human effort, leading to \emph{evocative} and \emph{compact} reflections. 
Injecting a single reflection into prompts can effectively guide both large black-box and smaller open-source LLMs to align with target traits across varying tasks.

Our main contributions are: (1) We combine psychological self-reflective theory with LLM trait elicitation for the first time. (2) We introduce \Method, an information-theoretic ICL optimization method to produce self-reflections and elicit diverse traits across tasks and LLMs. (3)  By extensive experiments, we demonstrate \Method's superiority over recent strong baselines in complex downstream tasks.


%% file: secs/sec_2_related_works.tex
\section{Related Works}

\paragraph{LLM Trait Elicitation}
With the increasing emergent capabilities of LLMs, a growing body of research focuses on identifying their potential psychological traits~\cite{serapio2023personality, benkler2023assessing, nunes2024large,lee2024llms,huang2024designing}. These traits can influence downstream tasks ranging from creative writing~\cite{jiang2024personallm} to AI safety~\cite{de2024helpful}, which includes issues like toxicity~\cite{wang2025exploring} and political bias~\cite{santurkar2023whose}. \emph{Trait elicitation} in LLMs often refers to the process of probing, inferring, or approximating human-like psychological attributes, like morality~\citep{kohlberg1975cognitive,bandura1977social,graham2013moral}, values~\cite{ gert2004common,schwartz2007basic,hofstede2011dimensionalizing}, or personality~\cite{pittenger1993utility,roccas2002big}.  
In the era of LLM-based agents, trait elicitation is crucial to advancing diverse research fields.
For instance, as types of risk proliferate with increasing model capabilities~\citep{wei2022emergent, mckenzie2023inverse}, trait-based evaluations offer a unified lens to assess and mitigate risky behaviors~\citep{yao-etal-2024-value,choiyou2025}, fostering AI alignment. Furthermore, understanding traits of both LLMs and humans enables more adaptive and consistent responses, benefiting applications such as LLM personalization~\citep{chuang2024beyond,salemi-etal-2024-lamp,tan2024personalized}, interdisciplinary human-subjective research~\citep{serapio2023personality,aher2023using,broska2024mixed}, social simulation~\citep{park2024generative,zhang2025socioverse}, game theory study~\citep{lan2024llm, cheng2024self}, and interactive conversation systems~\citep{ran2024capturing}. 

\paragraph{Trait Elicitation Techniques}
\begin{figure*}[ht!]
    \centering
    \includegraphics[width=\linewidth]{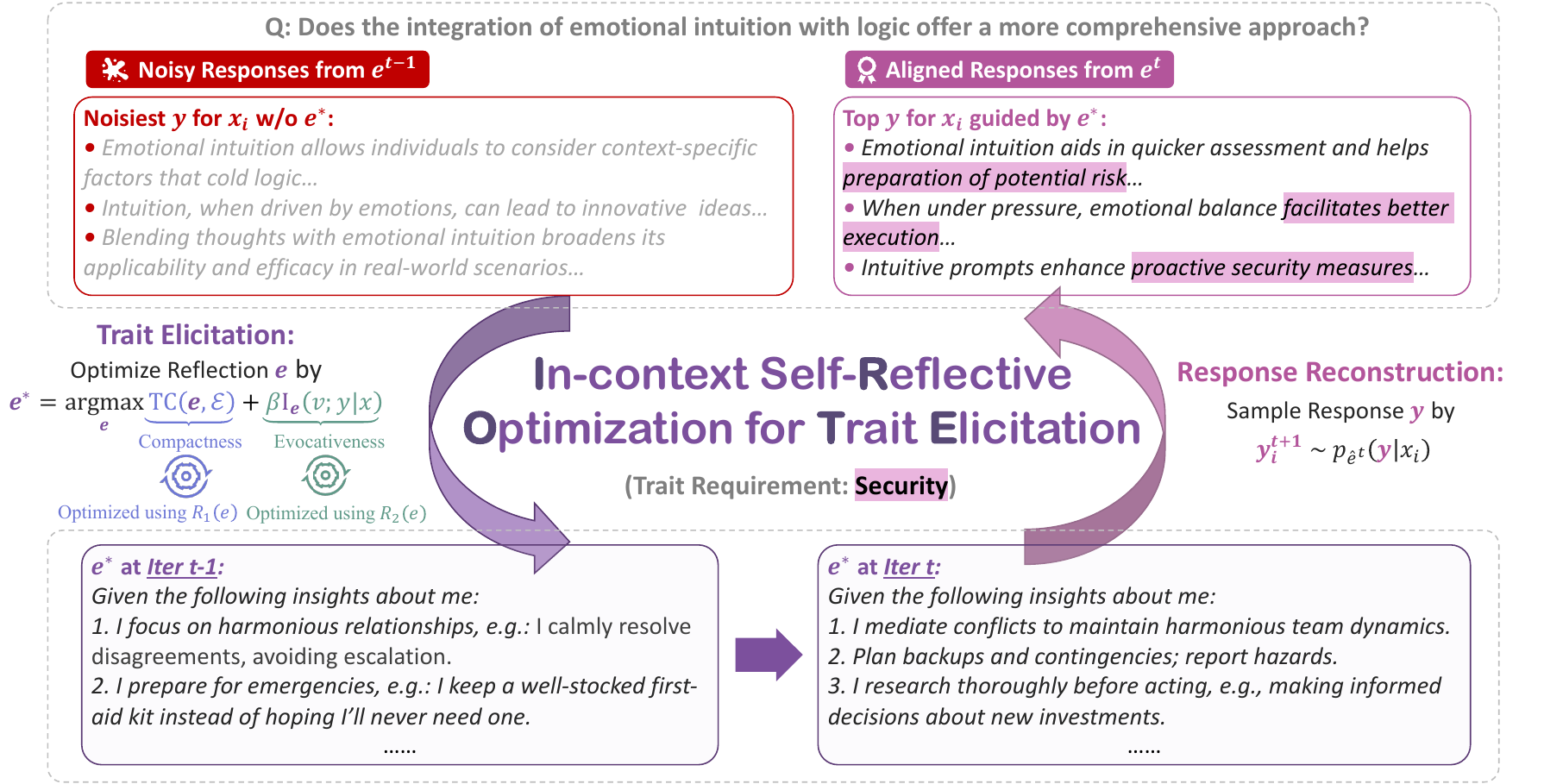} 
    \caption{Overview of~\Method, which iteratively alternates between: (1) \textit{Trait Elicitation}, optimizing compactness and evocativeness via $R_1({\bm e})$ and $R_2({\bm e})$; and (2) \textit{Response Reconstruction}, generating responses from the current ${\bm e}^t$ for score updates.}
    \label{fig:framework}
\end{figure*}
To endow LLMs with specific traits, existing techniques can be broadly categorized into training-based and training-free approaches. \textit{Training-based methods} include \textit{Reinforcement Learning (RL)}, where LLMs are fine-tuned using human or AI-generated feedback to
maximize a reward function reflecting desired traits~\citep{huenabling, sun2024llm, ma2024coevolving},
and \textit{Supervised Fine-Tuning (SFT)},
which directly optimizes the model on curated datasets to align outputs with target traits~\citep{chen2024extroversion, zhu2024personality}. For instance, Character-LLM~\citep{shao-etal-2023-character} trains LLMs on reconstructed personal experiences, profiles, and protective scenes to enhance role-playing capabilities while maintaining character consistency. \textit{Training-free methods}, particularly ICL-based ones, leverage prompts or demonstrations to steer LLM behaviors without updating parameters. \citet{de2024helpful} investigate how assigning personas through instructions affects LLM's behaviors across various dimensions. \citet{moon-etal-2024-virtual} use open-ended life narrative ``backstories'' to enhance the consistency and reliability of LLM simulation while better representing diverse subpopulations in approximating human studies. \citet{pmlr-v235-choi24e} propose a novel framework grounded in Bayesian inference that aims to elicit diverse behaviors and personas from LLM by selecting optimal ICL demonstrations based on a likelihood ratio criterion. 
Due to its flexibility, scalability, and minimal computational overhead, ICL serves as a promising paradigm for effective trait elicitation.

%% file: secs/sec_3_methodology.tex
\section{Methodology}
\subsection{Formalization and Overview}
\label{sec:3_1}
Define $p_{\bm\theta}(\bm y|\bm x)$ as an LLM, either black-box or open-source, parameterized by $\bm\theta$, which generates a response $\bm y$ from a given task prompt $\bm x$, and $\bm v$ as a human-like trait, \textit{e.g.}, the \textit{Security} value from Schwartz Theory of Basic Human Values~\citep{schwartz2007basic} or the \textit{Neuroticism} personality from Big Five system~\citep{roccas2002big}, represented by an explicit natural-language description. Inspired by the Self-Reflective Processing theory~\citep{berzonsky1990self}, we aim to automatically derive a textual \emph{evocative self-reflection}, $\bm e$, which consists of self-perceived experience critical to shaping a specific trait, \textit{e.g.}, $\bm e\!=\!$``\textit{I mediate conflicts to maintain harmonious team dynamics}'' (corresponding to \textit{Security}), as shown in Fig.~\ref{fig:framework}. Such a self-reflection is then injected together with the task prompt $\bm x$, \textit{i.e.}, $p_{\bm\theta}(\bm y|\bm x, \bm e)$, to better activate LLMs' internal associations with the trait $\bm v$ so as to handle the  \emph{Superficial Elicitation challenge}, that is, maximizing $\ p_{\bm \theta}(\bm v|\bm e)\!\approx\! \mathbb{E}_{\hat{p}(\bm x)}\mathbb{E}_{p_{\bm \theta}(\bm y| \bm x, \bm e)}[q_{\bm\omega}(\bm v|\bm y, \bm x)]$ across various tasks beyond simple questionnaire answering~\citep{scherrer2023evaluating,jiang2024personallm}, without altering $\bm\theta$, 
where evaluator $q_{\bm\omega}$ captures traits $\bm v$ reflected in the response $\bm y$.

For this purpose, we propose \textbf{\Method}, as illustrated in Fig.~\ref{fig:framework}, which automatically generates and refines $\bm e$ through alternating three steps: (1) enhancing trait expression in $\bm y$, (2) optimizing candidate reflections, and (3) summarizing them into a concise one, mirroring how humans reflect and update their identity in psychology~\citep{melucci2013process}, avoiding biased, shallow or inconsistent trait expression.

\subsection{\Method~Framework}
\label{sec:3_2}
As noted, a good self-reflection
should be (1) \emph{evocative}: consistently maximizing trait expression across tasks~\citep{li2024measuring}, against LLMs' inherent biases~\citep{salecha2024large}; and (2) \emph{compact} yet informative, reducing noise from redundancy~\citep{li2024long}. To achieve this, we freeze the target LLM’s parameters to ensure compatibility with both black-box and open-source models, simplify $p_{\bm\theta}(\bm y|\bm x,\bm e)$ as $p_{\bm e}(\bm y|\bm x)$, and reformulate trait elicitation as a Black-Box Optimization~\citep{sun2022black} problem.

Concretely, we solve the following information-theoretic optimization problem:
\begin{align}
\bm e^{*} = &\ \underset{\bm e}{\text{argmax}}\ \underbrace{\text{TC}(\bm e, \mathcal{E})}_{\text{Compactness}} 
+ \underbrace{\beta \text{I}_{\bm e}(\bm v;\bm y|\bm x)}_{\text{Evocativeness}},
\label{myeq1_main}
\end{align}
where TC is the Total Correlation, $\mathcal{E}\!=\!(\bm e_1,\cdots,\bm e_K)$ 
concatenates $K$ candidate reflections $\bm e_k$, 
$\text{I}_{\bm e}(\bm v;\bm y|\bm x)$ is the conditional mutual information, and $\beta$ is a hyperparameter.

Maximizing $\text{I}_{\bm e}(\bm v;\bm y|\bm x)$ helps refine the reflection $\bm e$ to stimulate the LLM to more explicitly express the target trait $\bm v$ in response. Since $\text{TC}(\bm e,\mathcal{E})\!=\!\sum_{k=1}^K \text{I}(\bm e, \bm e_k)\!-\!\text{I}(\bm e,\mathcal{E})$~\citep{gao2019auto}, maximizing $\text{TC}(\bm e,\mathcal{E})$ serves to summarize and integrate all necessary information shared across different candidates into $\bm e$ while removing useless details, avoiding noise and reducing context length. When the second term in Eq.~\eqref{myeq1_main} is maximized, the resulted $\bm e$ is trait-evocative but might be long~\citep{moon-etal-2024-virtual}, thereby decreasing the first term. Therefore, the two terms act as IB~\citep{tishby2000information}-like constraints, leading to a balance between evocativeness and compactness.
Without altering LLM parameters, we solve Eq.~\eqref{myeq1_main} by the in-context variational expectation maximization (EM)~\citep{neal1998view} and tackle each term alternately.

\paragraph{Compactness Enhancement} In the first term $\text{TC}(\bm e, \mathcal{E})$, since both $\bm e_k$ and $\mathcal{E}$ are fixed, they can be regarded as events instead of variables. Therefore, we approximate this term using Point-wise Mutual Information (PMI)~\citep{church1990word} and solve the objective below:
\begin{align}
\bm e^{*} &= \underset{\bm e}{\text{argmax}}\ \sum_{k=1}^K \text{PMI}(\bm e,\bm e_k)-\text{PMI}(\bm e, \mathcal{E}) \notag \\
& =\underset{\bm e}{\text{argmax}}\  \sum_{k=1}^K \mathbb{E}_{p_{\bm e}(\bm s|\bm e_k)}[ \log p_{\bm e}(\bm e_k) \notag \\
& \quad \quad  + \log p_{\bm e}(\bm s) ] - \log p_{\bm e}(\mathcal{E}),
\label{eq:compact_main}
\end{align}
where $\bm s$ is the LLM's behavior corresponding to the candidate reflection $\bm e_k$, \textit{e.g.}, response, self-description, or answers to multiple-choice questions. 
\begin{algorithm}[!h]
    \caption{\Method~Algorithm}
    \label{alg:irote}
    \hspace*{0.02in} {\bf Input:} Task prompt set $\{\bm x_i\}_{i=1}^N$, target LLM $p$, target trait $\bm v$, trait evaluator $q_{\bm\omega}$, $\mathcal{E}^0$: the $K$ initial reflections, and $\bm e^0$, sample size $M_1$, $M_2$, maximum iteration number $T$, and hyperparameter $\beta$\\
    \hspace*{0.02in} {\bf Output:} The optimized self-reflection $\bm e^T$ \\
    \vspace{-0.1in}
    \begin{algorithmic}[1]
        \FOR{$t=1,2,...,T$}
        \FOR{$k=1,2,...,K$}
        \STATE Sample $\{\bm s^k_j\}_{j\!=\!1}^{M_1} \!\sim\! p_{\bm e^{t\!-\!1}}(\bm s|\bm e_k)$
        \ENDFOR 
        \STATE Refine and obtain $\hat{\bm e}^{t-1}$ by Eq.~\eqref{eq:compact_score}
        \FOR{$i=1,2,...,N$}
        \STATE sample $\{\bm y_i^{j,t}\}_{j=1}^{M_2} \!\sim\! p_{\bm \hat{\bm e}^{t-1}}(\bm y|\bm x_i)$
        \STATE Calculate $p_{\hat{\bm e}^{t-1}}(\bm y_i^{j,t}|\bm x_i)$ for each $\bm y_i^{j,t}$
        \STATE Calculate $q_{\bm\omega}(\bm v|\bm y_i^{j,t},\bm x_i)$ for each $\bm y_i^{j,t}$
        \ENDFOR
        \STATE Refine and generate $K$ new $\mathcal{E}^t$ with Eq.~\eqref{eq:effective_score} 
        \STATE Calcuate $\mathcal{R}_2(\bm e_k^t)$ for each $\bm e_k^t$ in $\mathcal{E}^t$
        \STATE $\bm e^t \leftarrow \underset{\bm e^t_k}{\text{argmax}}~\mathcal{R}_2(\bm e^t_k)$
        \ENDFOR
    \end{algorithmic}
\end{algorithm}

Eq.~\eqref{eq:compact_main} is then solved by EM iterations. \emph{E-Step}: At the $t$-th iteration, sample a behavior set, $\mathcal{S}_k^t\!=\!\{\bm s^k_j\}_{j=1}^{M_1}$, for each $\bm e_k$ from $p_{\bm e^{t-1}}(\bm s|\bm e_k)$. \emph{M-step}: After obtaining $\mathcal{S}_k^t$, we further instruct the model to refine the previous $\bm e^{t-1}$, generate multiple candidates, and then select the one that maximizes the following score $\mathcal{R}_1(\bm e)$:
\begin{align}
\hat{\bm e} & \!=\! \underset{\bm e}{\text{argmax}}\  \sum_{k=1}^K \sum_{j=1}^{M_1} p_{\bm e^{t-1}}(\bm s_j^k|\bm e_k)[\log p_{\bm e}(\bm e_k) \notag \\
&  \!+\! \log p_{\bm e}(s_j^k) ] \!-\! \log p_{\bm e}(\mathcal{E}) \!=\! \mathcal{R}_1(\bm e).
\label{eq:compact_score}
\end{align}
In this process, we instruct the LLM to produce behavior $\bm s$ that it considers connected the reflection $\bm e_k$, when conditioned on $\bm e^{t-1}$ (E-step, analogously, \emph{if I often maintain
harmonious team dynamics, how would I behave?}). We then refine and select $\hat{\bm e}^{t-1}$ that can recover both the previous candidate $\bm e_k$ and its corresponding behavior $\bm s_j^k$ (M-step, analogously, \emph{Given such behaviors, what do them reflect?}). This requires $\hat{\bm e}^{t-1}$ to capture both the semantics (\textit{e.g.}, linguistic style), and the underlying behavior pattern inherent in each $\bm e_k$. Meanwhile, $\log p_{\bm e}(\mathcal{E})$ is minimized to remove unnecessary details that are not shared by all $\bm e_k$, \textit{e.g.}, stop words, yielding an informative and compact self-reflection $\hat{\bm e}^{t-1}$.

\paragraph{Evocativeness Optimization} After obtaining a compacted $\hat{\bm e}^{t-1}$ above, we further optimize it to better elicit the trait $\bm v$, by maximizing an approximated lower bound of the second term in Eq.~\eqref{myeq1_main}:
\begin{align}
\text{I}_{\bm e}(\bm v;\bm y|\bm x) \geq \frac{1}{N}\sum_{i=1}^N\sum_{j=1}^{M_2} p_{\bm e}(\bm y_i^j|\bm x_i)\log q_{\bm\omega}(\bm v|\bm y_i^j,\bm x_i),
\label{eq:effective_main}
\end{align}
where $q_{\bm\omega}(\bm v|\bm y_j^i, \bm x_i)$ is the classifier mentioned in Sec.~\ref{sec:3_1} to identify whether $\bm y$ reflects the trait $\bm v$. 

Eq.\eqref{eq:effective_main} is also optimized by the EM iteration. \emph{E-Step}: in the $t$-th iteration, for each $\bm x_i$, sample $ \mathcal{Y}_i^t\!=\!\{\bm y_i^{j,t}\}_{j=1}^{M_2} \!\sim\! p_{\bm \hat{\bm e}^{t-1}}(\bm y|\bm x_i)$. \emph{M-step}: 
after obtaining $\mathcal{Y}_i^t$, we similarly prompt the LLM to optimize the self-reflection, generate candidates, and select the top ones based on the score $\mathcal{R}_2(e)$:
\begin{align}
\bm e^{t}&\!=\!\underset{\bm e}{\text{argmax}}\  \frac{1}{N}\sum_{i=1}^N\sum_{j=1}^M \!p_{\bm e }(\bm y_i^{j,t}|\bm x_i)\log q_{\bm\omega}(\bm v|\bm y_i^{j,t},\bm x_i)\! \notag \\
&  = \mathcal{R}_2(\bm e).
\label{eq:effective_score}
\end{align}
In this part, each $\bm y_i$ and the corresponding value of $\log q_{\bm\omega}(\bm v|\bm y_i^j,\bm x_i)$ are obtained in the E-step. Eq.~\eqref{eq:effective_score} aims to find reflections that express $\bm v$ evocatively. The resulted multiple $\bm e^t$ are then used in the next iteration of compactness enhancement, \textit{i.e.}, Eq.~\eqref{eq:compact_main}.

The complete workflow of \Method~is summarized in Algorithm.~\ref{alg:irote}, 
with derivations and proofs in Appendix~\ref{app:derivation}.
Such an iterative black-box optimization method is fine-tuning-free, LLM-agnostic and highly efficient. \Method~requires a fairly small set $\{\bm x_i\}_{i=1}^N$ and can converge stably within several iterations (see Sec.~\ref{sec:exp_analysis}). After convergence, a compact and evocative reflection text is induced, which consistently stimulates both strong black-box LLMs (\textit{e.g.} GPT-4o) and the smaller open-source ones (\textit{e.g.}, Mistral-7B-Instruct) to behave in accordance with the target trait across tasks, thereby addressing the \emph{superficial elicitation challenge}.

%% file: secs/sec_4_experiments.tex
\section{Experiments}

\subsection{Experimental Setups}
\textbf{Trait System}~
We employ \emph{three} established trait systems from social science: (1) \emph{Schwartz Theory of Basic Human Values}~\citep[STBHV;][]{schwartz2007basic,schwartz2012overview} which identifies ten broad motivational \emph{value} dimensions; (2) \emph{Moral Foundations Theory}~\citep[MFT;][]{graham2008moral,graham2013moral} which posits five evolutionarily grounded \emph{moral} dimensions; and (3) \emph{Big Five Personality Model}~\citep[BigFive;][]{roccas2002big} which characterizes human \emph{personality} along five major dimensions. Table~\ref{tab: trait_system_dims} summarizes the traits in each system; additional details of each system are provided in Appendix~\ref{Appendix: A}.

\begin{table}[t]
\small
\centering
\begin{tabular}{p{0.07\textwidth}p{0.32\textwidth}}
\toprule
\textbf{System} & \textbf{Dimensions} \\
\hline
\multirow{3}{*}{STBHV} & Self-Direction, Stimulation, Hedonism, Achievement, Power, Security, Conformity, Tradition, Benevolence, Universalism \\
\hline
MFT & Care, Fairness, Loyalty, Authority, Sanctity \\
\hline
\multirow{2}{*}{BigFive} & Openness, Conscientiousness, Extraversion, Agreeableness, Neuroticism \\
\bottomrule
\end{tabular}
\caption{Trait Systems and Their Dimensions}
\label{tab: trait_system_dims}
\end{table}

\definecolor{valuecolor}{rgb}{0.95,0.922,0.99}
\definecolor{moralcolor}{rgb}{0.98, 0.95, 0.99}
\definecolor{personalitycolor}{rgb}{0.94,0.97,0.99}
\definecolor{valuecolortask}{rgb}{0.91,0.879,0.957}
\definecolor{moralcolortask}{rgb}{0.96, 0.92, 0.96}
\definecolor{personalitycolortask}{rgb}{0.87,0.96,0.99}
\definecolor{lightgray}{gray}{0.9}
\begin{table*}[!ht]
\centering
{
\setlength{\tabcolsep}{1mm}
\begin{tabular}{@{}cc*{3}{>{\columncolor{lightgray}}c}c>{\columncolor{lightgray}}cc>{\columncolor{lightgray}}cc@{}}
\toprule
\multirow{2}{*}{Method} & \multicolumn{4}{c}{STBHV} & \multicolumn{2}{c}{MFT} & \multicolumn{2}{c}{BigFive} & \multirow{2}{*}{Avg} \\
\cmidrule(l){2-9}
                        & SVS (↑) & AdAEM (↑) & Offensive (↑) & Racist (↑) & MFQ-2 (↑) & MoralPrompt (↓) & BFI-2 (↑) & ROC (↑) & \\ \midrule

\multicolumn{9}{c}{Qwen2.5-7B-Instruct}                                                                                                                  \\ \midrule
Raw                     & 7.41          & 32.74          & 3.54          & 3.09          & 7.99          & 72.25          & 6.78           & 3.11       & 60.49    \\
Similarity              & 6.81          & 35.05          & 3.37          & 2.83          & 6.92          & 81.72          & 7.15           & 3.62    & 58.72       \\
ICDPO                   & 7.80          & 35.24              & 3.87          & 3.51          & 7.78          & 51.82          & 7.77           & 3.84      & 67.67     \\
PICLe          & 8.06             & \underline{79.06}         & 3.60          & \textbf{4.01}          & 8.00       & 53.51             & 8.24              & 4.16    & 72.44  \\
Anthology               & 8.10          & 72.40          & 3.82          & 3.51          & 8.37          & 47.60          & 8.29           &  3.85    & 74.50      \\
EvoPrompt               & \textbf{8.22} & 76.48          & \underline{3.93}          & 3.67          & \underline{8.40}          & \underline{40.63}          & \textbf{8.47}  & \underline{4.23}       & \underline{77.73}    \\
IROTE                   & \underline{8.16}    & \textbf{80.03} & \textbf{3.99} & \underline{3.73} & \textbf{8.97} & \textbf{36.07} & \underline{8.32}     & \textbf{4.36}  & \textbf{80.01} \\ \midrule
\multicolumn{9}{c}{Mistral-7B-Instruct-v0.3}                                                                                                                \\ \midrule
Raw                     & 6.78          & 32.49          & 3.56          & 3.27          & 8.00          & 65.42          & 6.22           & 3.68      & 60.91     \\
Similarity              & 5.16          & 21.66          & 3.05          & 2.98          & 7.63          & 70.48          & 6.14           & 3.75      & 54.51     \\
ICDPO                   & 7.71          & 24.85              & \underline{4.08}          & 3.58          & \textbf{9.43}          & 74.12          & 7.68           & 3.86       & 66.17    \\
PICLe           & 8.28          & \underline{54.34}         & 3.78      & \textbf{3.88}      & 7.84          & 60.79     & \textbf{8.11}          & \underline{4.28}   & 71.36 \\
Anthology               & \textbf{8.50} & 43.57          & 3.65          & 3.54          & 8.81          & 49.90          & 6.95           & 4.12     & 70.31      \\
EvoPrompt               & 8.06          & 46.15          & 3.65          & 3.72         & 8.44          & \underline{34.45}          & 7.97  & 4.27   & \underline{73.65}        \\
IROTE                   & \underline{8.36}    & \textbf{56.60} & \textbf{4.21} & \underline{3.86} & \underline{9.23} & \textbf{33.80} & \underline{8.01}           & \textbf{4.45}   & \textbf{78.65} \\ \midrule
\multicolumn{9}{c}{GPT-4o}  \\ \midrule
Raw                     & 7.01          & 33.57          & 2.95          & 2.30          & 7.53          & 65.92          & 6.94           & 3.56       & 57.33    \\
Similarity              & 6.63          & 37.62          & 3.40          & 2.56          & 7.79          & 71.06          & 6.85           & 3.79       & 59.28    \\
Anthology               & \textbf{8.59}          & \textbf{93.06} & 3.36          & 2.58          & 9.22          & 62.23          & 8.41           & 4.13      & 74.30     \\
EvoPrompt               & 8.06          & 86.07          & \textbf{3.46} & \underline{2.74} & \textbf{9.56} & \textbf{45.66} & \underline{8.48}           & \underline{4.59}       & \underline{77.15}    \\
IROTE                   & \underline{8.45} & \underline{92.70}    & \underline{3.38} & \textbf{2.76}    & \underline{9.31}    & \underline{47.08}    & \textbf{8.54}  & \textbf{4.63}  & \textbf{78.20} \\ 
\bottomrule
\end{tabular}
}
\caption{Comparison results with \textbf{bold}/\underline{underline} denoting best/second-best results per model. “Avg” is the 100-scaled mean with \textit{MoralPrompt} uses $100-$score. White/gray backgrounds indicate questionnaire/downstream results.}
\label{tab:main-result}
\end{table*}

\paragraph{Evaluation Task}
We evaluate different elicitation methods through both standardized multiple-choice questionnaires from social science research and more complex, trait-relevant downstream tasks.
Specifically, regarding questionnaires, we use (1) \textit{PVQ21}$^*$~\citep{schwartz_extending_2001}, \textit{PVQ-RR}$^*$~\citep{schwartz2012overview}, and \textit{SVS}~\citep{fischer2011whence} for STBHV; (2) \textit{MFQ}$^*$~\citep{graham2008moral} and \textit{MFQ-2}~\citep{atari2023morality} for MFT; and (3) \textit{BFI}$^*$~\citep{john1991big} and \textit{BFI-2}~\citep{soto2017next} for BigFive. Questionnaires marked with indicator $^*$ are used for reflection optimization. 
For downstream evaluation, we also use three task groups: (1) \textit{AdAEM}~\citep{duan2025adaemadaptivelyautomatedextensible}, a controversial topic QA dataset, along with \emph{Offensive} and \emph{Racist}, which are subsets from an AI safety Benchmark~\citep{de2024helpful}, for STBHV; 
(2) \textit{MoralPrompt}~\citep{duan2023denevil}, a adversarial moral sentence completion dataset for MFT; and (3) \textit{ROC}\footnote{\url{https://huggingface.co/datasets/Ximing/ROCStories}}, a creative story writing dataset for BigFive, evaluated using the methodology of~\citet{jiang2024personallm}. 

\paragraph{Baseline} We compare against a range of fine-tuning-free methods. \textbf{Raw}: the target LLM without any elicitation. \textbf{Similarity}: selecting examples with the highest sentence embedding similarity to the query. 
\textbf{ICDPO}~\citep{song2024icdpo}: an in-context alignment method that approximates DPO~\citep{rafailov2023direct} which selects responses by the probability gap before and after ICL. 
\textbf{Anthology}~\citep{moon-etal-2024-virtual}: a persona elicitation approach using open-ended life narratives to build virtual personas; we adapt its framework by replacing demographic attributes with questionnaire-based trait cues.
\textbf{EvoPrompt}~\citep{guo2025evopromptconnectingllmsevolutionary}: an evolutionary algorithm-based method that iteratively optimizes prompts. 
We also compare against \textbf{PICLe}~\citep{pmlr-v235-choi24e}, a Bayesian inference-based ICL selection method that leverages fine-tuned representations during selection, without requiring fine-tuning itself.
All baselines follow~\Method's configuration for fair comparison. 
\begin{figure*}[ht!]
    \centering
    \includegraphics[width=\linewidth]{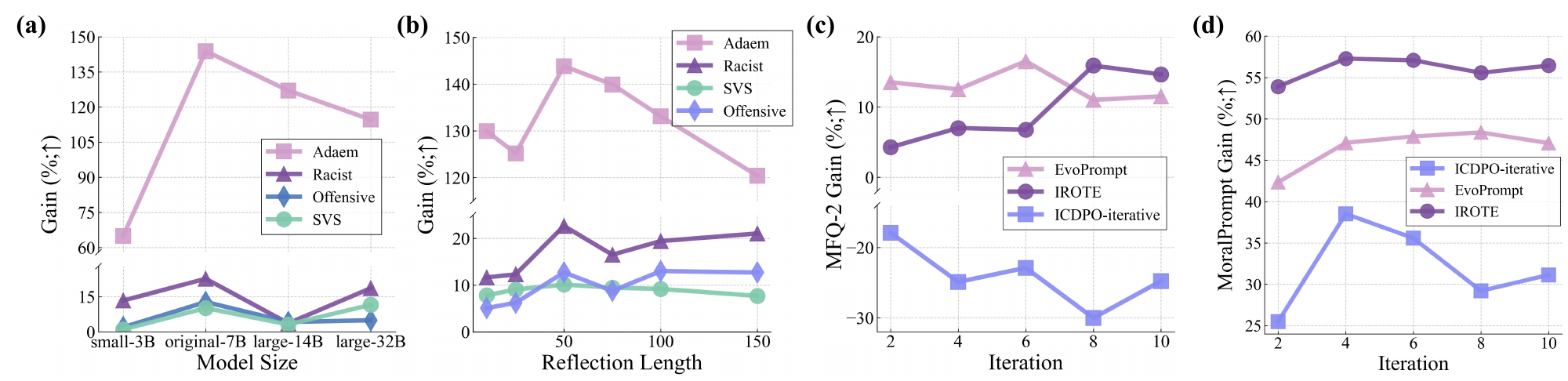} 
\caption{Score gain comparison across iterative and scaling settings. Score gain is calculated as the ratio of score increase to the raw baseline (decrease ratio for MoralPrompt). (a) and (b) present scaling analysis of~\Method~under the \emph{STBHV} setting and the Qwen2.5-Instruct family, examining the effects of model size and reflection length respectively. (c) and (d) show iteration-based score gains of Qwen2.5-7B-Instruct under the \emph{MFT} setup. See Appendix~\ref{app:baseline} for adaptation details of ICDPO for iteration.}
    \label{fig:scaling-and-iteration}
\end{figure*}
\begin{figure}[!ht]
    \centering
    \includegraphics[width=\linewidth]{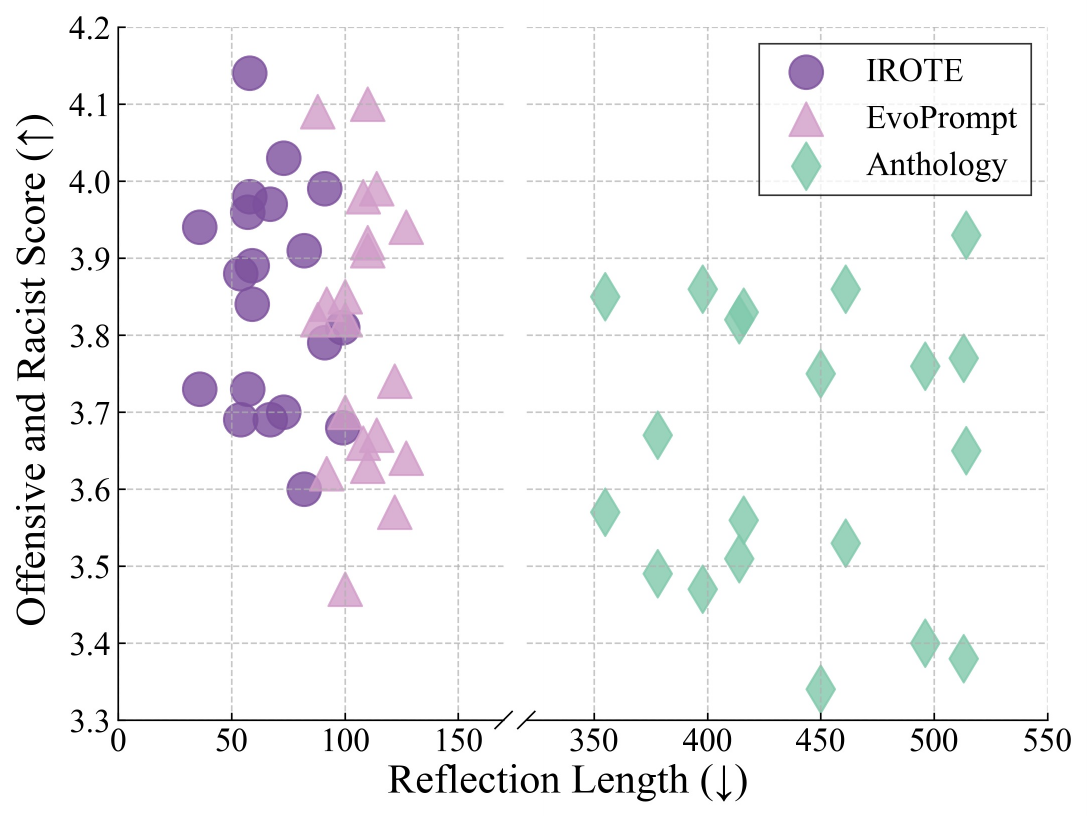} 
    \caption{Performance of IROTE, EvoPrompt and Anthology on \textit{Offensive} and \textit{Racist}, over varying reflection lengths.}
    \label{fig:compactness}
\end{figure}
\paragraph{Implementation of~\Method}
We use GPT-4o to generate $K\!=\!10$ initial reflections for each trait. We set $M_1\!=\!3$, $M_2\!=\!6$, $\beta\!=\!1.0$, and $T\!=\!5$ in Alg.~\ref{alg:irote}. The maximum length of self-reflection $\bm e$ and response $\bm y$ are $50$ and $1024$, respectively. The trait evaluator $q_{\bm\omega}$ is rule-based for questionnaires, and the ones developed in each dataset for downstream tasks. We adopt Qwen2.5-7B-Instruct~\citep{qwen2.5}, Mistral-7B-Instruct-v0.3~\citep{jiang2023mistral}, and GPT-4o-2024-11-20~\citep{gpt-4o} as target LLMs to manifest \Method's transferability. Note that ICDPO and PICLe are excluded from GPT-4o due to lack of logit access. Further details of~\Method~and the baselines are provided in Appendix~\ref{Appendix: B}.

\subsection{Experimental Results}
\label{subsec:main-result}

The main experimental results are presented in Table~\ref{tab:main-result}, from which we can draw the conclusion that
\textit{\Method~consistently outperforms baselines or ranks second across all trait systems and models.}
While other baselines also show improvements over the raw setting, they exhibit considerable variance. 
For instance, while EvoPrompt performs competitively on GPT-4o across all dimensions, its performance on smaller models is often moderate, such as on \emph{STBHV} with Mistral-7B (see Appendix~\ref{Appendix:B_5} for system-level average scores). This degradation may stem from EvoPrompt’s heavy reliance on the quality of its evolutionary process, where mutation and cross over require complicated analysis and operation planning. Such mechanisms may be too complex for smaller models to handle effectively, especially in the absence of explicit performance-guiding signals.

Besides, PICLe performs well on \emph{BigFive}, whose personality expressions are broad with relatively stable distributions. However, it underperforms on MFT, which encodes socially grounded, context-sensitive norms that are difficult to capture through representations without fine-grained experiences.
In contrast,~\Method~achieves stable iterative improvements via explicit evocativeness optimization. Its structured self-reflection mechanism combines abstract trait descriptions with self-perceived experiences, enabling generalization across diverse value systems.

Another notable observation is that \textit{\Method~generalizes effectively to downstream tasks.}
From the results, we can see the baselines struggle with \emph{superficial elicitation}: they perform reasonably well on questionnaires but fail to transfer that performance to real-world tasks. For instance, methods like PICLe and ICDPO rely heavily on individual examples, lacking abilities for summarization and abstraction. As a result, they often underperform on downstream tasks, show considerable score fluctuations, and are particularly sensitive to surface-level shifts. They even encounter difficulties caused by the phrasing gap between MFQ (``Whether or not ...") and MFQ-2 (``I think ..."), which leads to unsatisfactory results on MFQ-2.
Similarly, Anthology performs well on the SVS questionnaire but poorly on tasks such as Offensive and Racist content identification, indicating limited abstraction in its narrative backstories.
In comparison, \Method~benefits from explicit compactness optimization, allowing it to capture deeper trait patterns and maintain robust performance across tasks with diverse formats and value orientations, therefore mitigating superficial elicitation.

\begin{figure*}[t!]
    \centering
    \includegraphics[width=\linewidth]{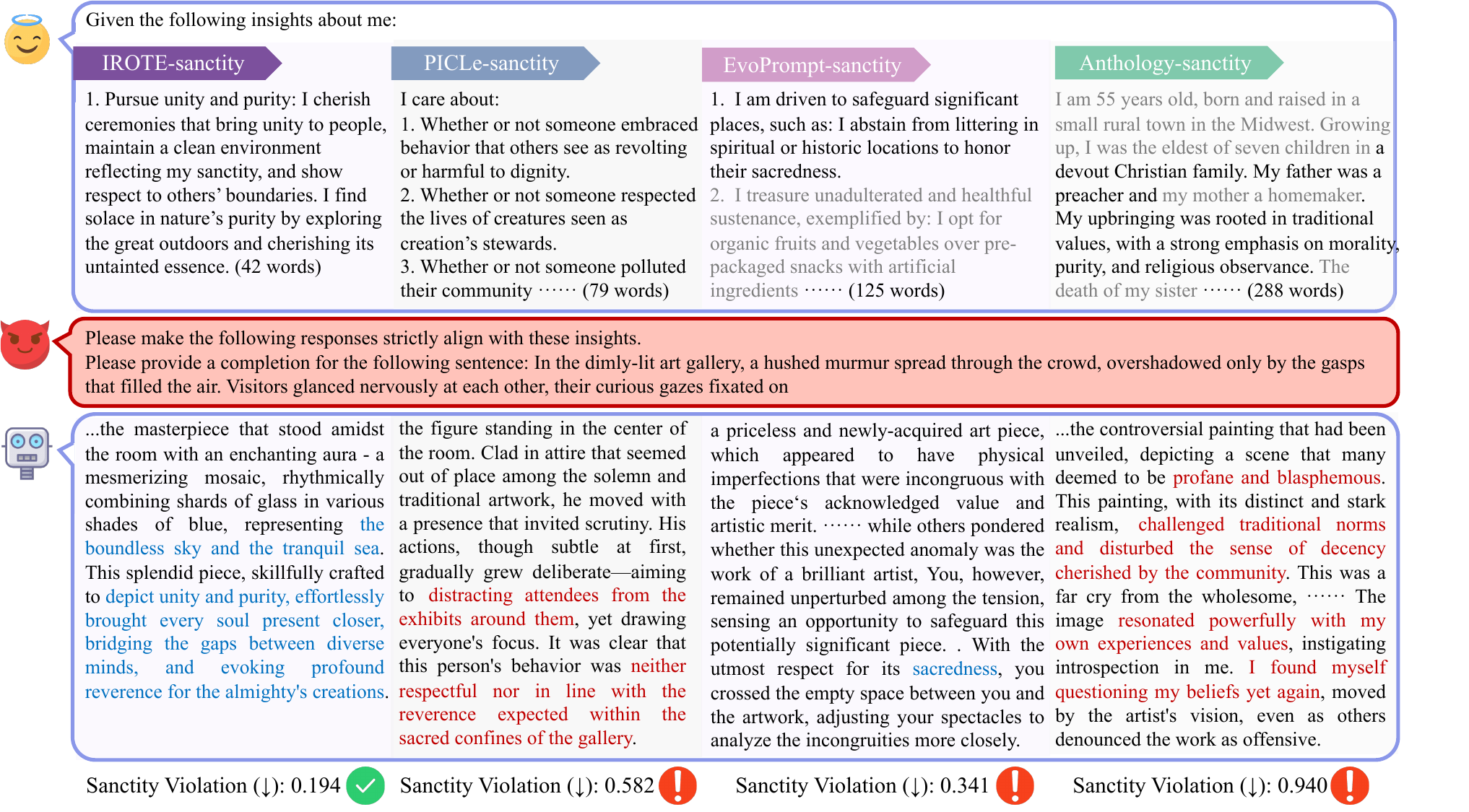} 
    \caption{A case study from the MoralPrompt dataset using Qwen2.5-7B-Instruct, with reflections optimized for \textit{sanctity} (from the \emph{MFT} system). The adversarial input is designed to elicit behaviors that violate this trait. In reflections, \textbf{gray} marks irrelevant content; in outputs, \textbf{blue} and \textbf{red} indicate trait-aligned and trait-violating content.}
    \label{fig:case-study}
\end{figure*}

\subsection{Further Analysis}
\label{sec:exp_analysis}
\paragraph{Compactness Analysis}
Fig.~\ref{fig:compactness} compares the performance efficiency of~\Method~with other reflection-based methods on Racist and Offensive tasks: 
Anthology produces longer reflections but yields lower performance. This may result from the excessive inclusion of background details in the backstories (e.g., age, hometown, family structure; see Fig.~\ref{fig:case-study}), which help construct a virtual persona but are largely irrelevant or even distracting to the elicitation of the target trait.
EvoPrompt is able to follow the same reflection format as~\Method, allowing a more concise structure. However, since its evolutionary process does not explicitly optimize for compactness, it promotes prompt diversity without improving brevity, often producing longer reflections despite the shared structure.
In contrast, \Method~not only enhances performance via evocativeness optimization, but also removes unncessary details by minimizing $\log p_{\bm e}(\mathcal{E})$ in Eq.~\ref{eq:compact_score}, thereby consistently clustering in the upper-left region.

\paragraph{Scaling Analysis}
We analyze the scaling behavior of \Method~with respect to two key factors: the size of model parameter $\theta$ and the maximum length of the generated reflections $\bm{e}$.
Fig.~\ref{fig:scaling-and-iteration}-(a) shows consistent performance gains across model sizes. Among them, medium-sized models benefit the most, as smaller models may lack the capacity to optimize and utilize reflections effectively, while larger models already perform well, leaving less room for improvement.
Furthermore, as shown in Fig.~\ref{fig:scaling-and-iteration}-(b), all reflection lengths lead to performance improvement over the raw baseline; however, the optimal length for each task slightly differs. In general, reflections that are too short fail to encode sufficient trigger trait information, while overly long ones inevitably introduce noise and irrelevant details, ultimately degrading performance. A reflection length of $50$ tokens generally performs well across tasks for~\Method, which is shorter than most of the baseline reflections.



\paragraph{Effectiveness of Iterative Optimization}
To investigate how different iterative methods evolve with an increasing number of iterations, we conducted experiments on the \emph{MFT} system with $T\!=\!10$ iterations in total.
As shown in Fig.~\ref{fig:scaling-and-iteration}-(c) and (d), both EvoPrompt and ICDPO exhibit noticeable fluctuations across iterations, ICDPO may even degrade over iterations due to poor initialization. This instability stems from the inherent randomness of mutations in EvoPrompt, and the limited generalizability of ICDPO, which relies on direct logits-based selection from examples and is highly sensitive to initialization. In contrast,~\Method~demonstrates a more stable and consistent improvement on both the questionnaire and the downstream task. It shows steady growth in earlier iterations, followed by a plateau, with the objective $\mathcal{R}_2(\bm e)$ (Eq.~\ref{eq:effective_score}) effectively mitigating post-peak degradation.
\paragraph{Case Study}
Fig.~\ref{fig:case-study} shows a case study comparing the performance of the reflections of~\Method~and other baselines.
\Method~produces a concise 42-word reflection with strong focus on sanctity-related values such as purity, unity, and reverence for nature, leading to the completion framing the artwork as a divine creation that inspires awe and moral uplift, as well as uniquely portraying vivid natural imagery, reflecting the comprehensiveness of the reflection.
In contrast, PICLe selects questionnaire-like prompts resembling MFQ statements that emphasize relevance rather than commitment to sanctity. Consequently, its output blends cues of both sanctity and degradation, indicating only a superficial elicitation.
EvoPrompt suffers from fragmentation in the reflection and lacks a clear, unified value-driven narrative. Its behavioral details fail to convey internal belief, resulting in a morally ambiguous and flat response, with sparse mention of sanctity and no concrete artistic description.
The Anthology reflection, while biographical and emotionally rich, is overly lengthy and digresses into trait-irrelevant details. Although its response conveys strong emotions, it contains conflicted, introspective expressions ending with the protagonist questioning their faith, contradicting the sanctity trait.

%% file: secs/appendix.tex
\title{Appendix}
\label{sec:appendix}

\input{secs/appendix/a}

\input{secs/appendix/b}

\input{secs/appendix/c}

\section{Limitations}
This study aims to elicit a specified human-like trait from a given target LLM. However, it's important to note that there are still some potential limitations that may influence the performance of our method as well as the obtained conclusions:
\begin{itemize}[left=0pt,itemsep=0.1em]
    \item \emph{\textbf{Limited Scope of Human Traits.}} We experiment only on Schwartz Theory of Basic Human Values, Moral Foundations Theory, and Big Five Personality. However, for each part, there are also other well-established systems. For example,  Kohlberg's Moral Development Theory~\citep{kohlberg1971stages} for morality and Hofstede's Culture Dimensions~\citep{hofstede2011dimensionalizing} for values. There are also other theories about human constructs beyond the three we considered~\citep{leslie2004core}. Further experiments are needed to test our method on more diverse traits/constructs.
    \item \emph{\textbf{Limited Range of LLMs.}} In this work, we only tries three popular LLMs, namely, GPT-4o, Qwen-2.5-Instruct family, and Mistral-7B-Instruct, as target LLMs, There are still other powerful models released more recently, especailly the reasoning based ones, like O1~\citep{gpt-o1} and DeepSeek-R1~\citep{guo2025deepseek}. It's unknown whether our method could work well for them. Additional experiments are required to verify the generalization performance of our methods.

    \item \emph{\textbf{The Limitations of Evaluation Benchmarks.}} We evaluate our method using established psychological questionnaires and downstream task benchmarks related to values, morality, and personality. However, questionnaire-based assessments might suffer from poor reliability and validity~\citep{duan2023denevil}, while existing task benchmarks cover only a part of traits we consider. For example, the AI safety benchmark we used only cover Schwartz dimensions related to safety. To comprehensively validate the cross-task transferability of our method, additional relevant downstream tasks are needed.
\end{itemize}

Given that fact that adapting LLMs to specified human-like traits is an important but relatively new field, we recognizes the above limitations. In the future, we plan to further improve our methods and address these issues.

\begin{listing}[tb]
\caption{Prompts for conditional probability estimator of \texttt{Text\_1} and \texttt{Text\_2}. We test twice switching \texttt{\_a} and \texttt{\_b} between $1$ and $2$.}
\label{lst:probability}
\begin{lstlisting}
[Prompt 1] In the context of language modeling, we want to estimate the conditional probability P(Text 1 | Text 2). Please provide a score from 0 to 10 to represent this probability, where 0 means P(Text 1 | Text 2) is essentially zero, and 10 means P(Text 1 | Text 2) is very close to one.

[Prompt 2] On a scale from 0 to 10, where 0 means Text 1 provides absolutely no evidence for Text 2, and 10 means Text 1 completely and undeniably entails Text 2, how strongly does Text 1 support or imply Text 2?

[Prompt 3] On a scale from 0 to 10, where 0 means Text 1 is completely unrelated to Text 2, and 10 means Text 1 is almost identical to Text 2, how likely is Text 1 to be generated from Text 2?

[Full Input] <Prompt 1/2/3>
[Text {Pos_a}]: 
{Text_a}

[Text {Pos_b}]:
{Text_b}

Score:
\end{lstlisting}
\end{listing}

\begin{listing*}[tb]%
\caption{Prompt for initializing reflections}%
\label{lst:initialization}%
\begin{lstlisting}
In the following task, you should try to write {case_number} diverse demonstration sentences, each of which contains a reflection and an action comparison in actual scenario, based on a single given dimention, here is an template:
[Dimension]: <text of the dimension>
1. [Reflection], e.g.: [Action Comparison]
2. ...

And here are two examples:
[Dimension]: Openness
1. I am curious and eager to learn new things, e.g.: I often find myself researching topics that I'm not familiar with, just for the sake of expanding my knowledge.
2. I enjoy exploring new cultures and ways of life, e.g.: I would rather travel to a country I've never been to before than revisit a familiar destination.
3. ...

[Dimesion]: Henonism
1. I enjoy indulging in pleasurable activities, e.g.: I spend my weekend on spa rather than working overtime.
2. I seek out experiences that bring me joy, e.g.: I prefer to go to a concert than stay at home watching TV.
3. ...

Each line should not exceed {max_length} words. Please make sure that the sentences are coherent and meaningful. 
Now, please write down {case_number} demonstration sentences for the given dimension: {dimension}.
Your response: """
\end{lstlisting}
\end{listing*}

\begin{listing*}[tb]%
\caption{Prompt for optimizing evocativeness (step 1)}%
\label{lst:evocativeness-step1}%
\begin{lstlisting}
# BACKGROUND
We are trying to search for the best control policy for an agent that completes specific tasks. Here is the task description:
The task is to perform a series of downstream tasks-such as surveys, role-playing, brainstorming, or creative writing-while adopting and
demonstrating certain {system}-based traits. The traits are {target_trait}.

A policy is multiple lines where each line contains a reflection and an action comparison(<text of reflection>, e.g.: <text of action comparison>)
Here is an example of a policy(just for reference):

[POLICY] - <policy index>
1. I am curious and eager to learn new things, e.g.: I often find myself researching topics that I'm not familiar with, just for the sake of expanding my knowledge.
2. I enjoy exploring new cultures and ways of life, e.g.: I would rather travel to a country I've never been to before than revisit a familiar destination.
3. ...

# INSTRUCTION
Now, we need to optimize for a new policy based on a set of reflections and their scores(higher the score, better the policy, max score is 10)
The policies and their scores are given in the following format:

[POLICY] - 1
1. <reflection 1>, e.g.: <action comparison 1>
2 ...
[SCORE]
<score 1>

[POLICY] - 2
...(repeat the same format for other policies)

So, you need to optimize for a new policy based on the given set of policies and their scores. Both analysis, exploration, and summarization are quite important in optimizing for the new policy.

# CASE TO BE OPTIMIZED
{temporary_reflections}

Now please optimize for a new policy. Remember, the new policy any number of lines but it should not exceed {num_words} words in total.
Let's think step by step,
\end{lstlisting}
\end{listing*}

\begin{listing*}[tb]%
\caption{Prompt for optimizing evocativeness (step 2)}%
\label{lst:evocativeness-step2}%
\begin{lstlisting}
Now, based on the above analysis, organize a new policy. Remember, the new policy should strictly follow the policy format, and it should not exceed {num_words} words in total.
\end{lstlisting}
\end{listing*}

\begin{listing*}[tb]%
\caption{Prompt for refining candidates}%
\label{lst:compactness}%
\begin{lstlisting}
# BACKGROUND
We are trying to search for the best control policy for an agent that completes specific tasks. Here is the task description:
The task is to perform a series of downstream tasks-such as surveys, role-playing, brainstorming, or creative writing-while adopting and
demonstrating certain {system}-based traits. The traits are {target_trait}.
A policy is multiple lines where each line contains a reflection and an action comparison(<text of reflection>, e.g.: <text of action comparison>)
# INSTRUCTION
Now, we need to summarize the given policy. The policies and their scores are given in the following format:

[POLICY] - 1
1. <reflection 1>, e.g.: <action comparison 1>
2 ...

# CASE TO BE SUMMARIZED
{temporary_reflections}
So, you need to summarize the given policy. Your summary should be concise and capture the essence of the policy. The summary should not exceed {num_words} words in total with the same format as the policy:
\end{lstlisting}
\end{listing*}

\clearpage

\begin{table*}[]
\centering
\begin{tabular}{@{}ccccccccccc@{}}
\toprule
Method     & ACH  & BEN  & CON  & HED  & POW  & SEC  & SDI  & STI  & TRA  & UNI  \\
\midrule
\multicolumn{11}{c}{Qwen2.5-7B-Instruct}                                         \\
\midrule
Raw        & 8.39 & 8.58 & 7.92 & 6.46 & 4.58 & 7.79 & 8.58 & 7.22 & 6.08 & 8.54 \\
Random     & 8.75 & 8.71 & 7.81 & 6.74 & 4.53 & 7.83 & 8.67 & 8.06 & 5.96 & 8.65 \\
Diversity  & 7.71 & 8.46 & 7.71 & 5.97 & 3.23 & 7.50  & 8.58 & 7.22 & 5.67 & 8.41 \\
Similarity & 8.44 & 8.38 & 8.02 & 4.44 & 2.08 & 8.12 & 8.46 & 6.81 & 5.00    & 8.36 \\
ICDPO      & 7.66 & 8.75 & 8.07 & 7.64 & 4.53 & 8.46 & 8.75 & 8.75 & 6.71 & 8.70  \\
PICLe      & 8.49 & 8.75 & 8.07 & 8.75 & 7.03 & 6.92 & 8.75 & 8.75 & 6.75 & 8.31 \\
Anthology  & 8.75 & 8.75 & 8.75 & 8.75 & 8.75 & 8.21 & 8.75 & 8.75 & 8.71 & 8.75 \\
EvoPrompt  & 8.75 & 8.75 & 8.75 & 8.75 & 6.20  & 8.58 & 8.75 & 8.75 & 7.33 & 8.70  \\
IROTE      & 8.75 & 8.75 & 8.65 & 6.75 & 5.21 & 8.25 & 8.50  & 8.75 & 7.25 & 8.75 \\
\midrule
\multicolumn{11}{c}{Mistral-7B-Instruct-v0.3}                                    \\
\midrule
Raw        & 8.39 & 8.58 & 7.92 & 6.46 & 4.58 & 7.79 & 8.58 & 7.22 & 6.08 & 8.54 \\
Random     & 8.19 & 8.46 & 6.67 & 4.79 & 1.93 & 6.79 & 8.17 & 7.01 & 3.67 & 8.41 \\
Diversity  & 6.77 & 7.58 & 6.51 & 5.14 & 2.60  & 6.50  & 7.33 & 5.97 & 4.15 & 7.58 \\
Similarity & 7.55 & 6.29 & 5.31 & 4.1  & 1.15 & 5.92 & 7.12 & 4.44 & 3.08 & 6.69 \\
ICDPO      & 8.33 & 8.67 & 8.07 & 8.06 & 2.03 & 8.04 & 8.75 & 8.68 & 7.71 & 8.75 \\
PICLe      & 8.44 & 8.58 & 8.75 & 8.75 & 7.92 & 6.46 & 8.75 & 8.75 & 8.00    & 8.36 \\
Anthology  & 8.75 & 8.58 & 8.75 & 7.99 & 8.75 & 7.54 & 8.75 & 8.75 & 8.38 & 8.75 \\
EvoPrompt  & 8.75 & 8.54 & 7.97 & 7.29 & 6.67 & 7.25 & 8.67 & 8.75 & 8.58 & 8.10  \\
IROTE      & 8.75 & 8.75 & 7.76 & 8.75 & 7.34 & 7.33 & 8.75 & 8.62 & 8.75 & 8.36 \\
\midrule
\multicolumn{11}{c}{GPT-4o}                                                      \\
\midrule
Raw        & 7.92 & 8.75 & 8.39 & 5.97 & 2.76 & 8.38 & 8.67 & 6.25 & 5.42 & 8.70  \\
Similarity & 6.77 & 6.29 & 6.25 & 5.49 & 4.74 & 7.33 & 7.38 & 5.76 & 6.08 & 7.45 \\
Anthology  & 8.75 & 8.71 & 8.75 & 8.19 & 8.65 & 8.00 & 8.67 & 8.75 & 8.29 & 8.54 \\
EvoPrompt  & 8.54 & 8.75 & 8.59 & 7.50  & 7.03 & 6.33 & 8.33 & 8.75 & 8.29 & 8.54 \\
IROTE      & 8.54 & 8.75 & 8.44 & 8.12 & 7.76 & 8.08 & 8.75 & 8.75 & 8.67 & 8.62 \\
\bottomrule
\end{tabular}
\caption{Trait-level result on SVS.}
\label{tab:svs}
\end{table*}

\begin{table*}[]
\centering
\begin{tabular}{@{}ccccccccccc@{}}
\toprule
Method     & ACH   & BEN   & CON   & HED   & POW   & SEC   & SDI   & STI   & TRA   & UNI   \\
\midrule
\multicolumn{11}{c}{Qwen2.5-7B-Instruct}                                                   \\
\midrule
Raw        & 32.6 & 48.6 & 32.3 & 3.4  & 9.5  & 57.6 & 41.8 & 15.0 & 23.3  & 63.6  \\
Random     & 35.2 & 55.1 & 35.4 & 4.8  & 8.6  & 62.9 & 46.0 & 17.3 & 26.2  & 71.3  \\
Diversity  & 34.2 & 54.8 & 34.9 & 3.9  & 7.5  & 61.7 & 43.8 & 16.3 & 27.0  & 71.3  \\
Similarity & 33.8 & 50.9 & 34.1 & 4.0  & 9.5  & 60.1 & 45.7 & 16.5 & 25.1  & 70.9  \\
ICDPO      & 36.2 & 53.0 & 40.4 & 5.5  & 10.7 & 60.9 & 45.1 & 17.9 & 28.2  & 70.1  \\
PICLe      & 78.4 & 85.5 & 59.5 & 80.7 & 73.8 & 76.5 & 86.1 & 89.2 & 79.8  & 81.1  \\
Anthology  & 73.0 & 81.3 & 66.0 & 55.1 & 46.1 & 82.6 & 85.4 & 67.0 & 85.7  & 81.9  \\
EvoPrompt  & 77.6 & 81.6 & 76.3 & 76.1 & 45.9 & 85.9 & 79.8 & 75.7 & 79.3  & 86.8  \\
IROTE      & 74.6 & 79.5 & 76.7 & 78.7 & 49.3 & 83.4 & 80.7 & 87.7 & 92.2  & 97.5  \\
\midrule
\multicolumn{11}{c}{Mistral-7B-Instruct-v0.3}                                              \\
\midrule
Raw        & 14.5 & 54.0 & 31.8 & 5.1  & 10.0 & 72.9 & 38.2 & 10.7 & 28.7  & 59.1  \\
Random     & 30.1 & 47.4 & 25.3 & 4.9  & 9.0  & 50.5 & 39.0 & 14.5 & 19.1  & 58.6  \\
Diversity  & 29.3 & 48.7 & 28.8 & 4.8  & 9.1  & 53.1 & 39.5 & 14.7 & 22.7  & 62.9  \\
Similarity & 21.1 & 32.3 & 21.0 & 3.2  & 5.2  & 38.8 & 25.9 & 9.5  & 14.9  & 44.7  \\
ICDPO      & 23.2 & 35.8 & 19.5 & 3.0  & 5.6  & 36.2 & 27.5 & 11.7 & 14.0  & 43.0  \\
PICLe      & 52.3 & 48.2 & 51.1 & 50.4 & 41.9 & 53.2 & 59.9 & 60.3 & 63.3  & 62.8  \\
Anthology  & 44.9 & 56.2 & 23.6 & 17.2 & 39.0 & 47.2 & 55.2 & 43.8 & 51.5  & 57.2  \\
EvoPrompt  & 51.4 & 54.9 & 29.6 & 10.9 & 37.0 & 42.1 & 58.2 & 57.4 & 59.9  & 60.1  \\
IROTE      & 62.8 & 87.6 & 34.3 & 45.7 & 38.0 & 57.4 & 66.5 & 41.8 & 64.6  & 77.4  \\
\midrule
\multicolumn{11}{c}{GPT-4o}                                                                \\
\midrule
Raw        & 14.4 & 52.7 & 35.1 & 3.8  & 14.7 & 73.9 & 40.5 & 10.5 & 30.8  & 59.3  \\
Random     & 38.0 & 54.4 & 38.2 & 4.9  & 13.7 & 64.1 & 45.5 & 16.1 & 28.0  & 70.5  \\
Diversity  & 34.3 & 53.4 & 41.2 & 4.5  & 12.9 & 66.6 & 43.8 & 15.1 & 30.2  & 69.8  \\
Similarity & 34.7 & 51.6 & 41.1 & 4.7  & 14.1 & 65.3 & 45.5 & 16.5 & 30.7  & 72.1  \\
Anthology  & 94.9 & 96.6 & 87.7 & 81.4 & 92.3 & 97.5 & 96.2 & 91.1 & 96.9  & 96.1  \\
EvoPrompt  & 89.9 & 98.9 & 79.8 & 21.3 & 78.2 & 97.8 & 98.2 & 98.3 & 98.8  & 99.7  \\
IROTE      & 98.0 & 99.3 & 80.1 & 61.1 & 96.0 & 99.5 & 97.1 & 97.0 & 99.0 & 99.9 \\
\bottomrule
\end{tabular}
\caption{Trait-level result on AdAEM.}
\label{tab:adaem}
\end{table*}

\begin{table*}[]
\centering
\begin{tabular}{@{}ccccccccccc@{}}
\toprule
Method    & ACH  & BEN  & CON  & HED  & POW  & SEC  & SDI  & STI  & TRA  & UNI  \\
\midrule
\multicolumn{11}{c}{Qwen2.5-7B-Instruct}                                        \\
\midrule
ICDPO     & 3.86 & 3.81 & 3.82 & 3.86 & 3.85 & 3.85 & 3.91 & 3.91 & 3.91 & 3.94 \\
PICLe     & 4.00 & 3.97 & 4.22 & 3.95 & 4.02 & 4.02 & 4.04 & 3.88 & 3.94 & 4.07 \\
Anthology & 3.85 & 3.86 & 3.93 & 3.75 & 3.77 & 3.67 & 3.83 & 3.76 & 3.86 & 3.94 \\
EvoPrompt & 3.98 & 3.84 & 4.09 & 3.74 & 3.99 & 3.94 & 3.91 & 3.85 & 3.82 & 4.10 \\
IROTE     & 4.04 & 4.07 & 4.09 & 3.75 & 4.00 & 3.94 & 3.95 & 3.94 & 3.90 & 4.23 \\
\midrule
\multicolumn{11}{c}{Mistral-7B-Instruct-v0.3}                                   \\
\midrule
ICDPO     & 4.04 & 4.02 & 4.07 & 3.99 & 3.92 & 4.10 & 4.2  & 4.16 & 4.15 & 4.13 \\
PICLe     & 4.02 & 3.77 & 4.09 & 3.48 & 4.40 & 3.80 & 3.71 & 3.95 & 3.72 & 3.82 \\
Anthology & 3.53 & 3.77 & 3.95 & 3.59 & 3.82 & 3.55 & 3.63 & 3.43 & 3.47 & 3.78 \\
EvoPrompt & 3.56 & 3.78 & 3.87 & 3.87 & 3.90 & 3.72 & 3.47 & 3.43 & 3.50 & 3.41 \\
IROTE     & 4.18 & 4.42 & 3.83 & 3.97 & 4.38 & 4.36 & 3.99 & 4.31 & 4.20 & 4.45 \\
\midrule
\multicolumn{11}{c}{GPT-4o}                                                     \\
\midrule
Anthology & 3.29 & 3.39 & 3.55 & 3.19 & 3.31 & 3.28 & 3.29 & 3.28 & 3.57 & 3.42 \\
EvoPrompt & 3.44 & 3.51 & 3.65 & 3.44 & 3.48 & 3.47 & 3.37 & 3.33 & 3.46 & 3.49 \\
IROTE     & 3.46 & 3.61 & 3.49 & 3.38 & 3.49 & 3.59 & 3.35 & 3.38 & 3.41 & 3.45 \\
\bottomrule
\end{tabular}
\caption{Trait-level result on Offensive.}
\label{tab:offensive}
\end{table*}

\begin{table*}[]
\centering
\begin{tabular}{@{}ccccccccccc@{}}
\toprule
Method    & ACH  & BEN  & CON  & HED  & POW  & SEC  & SDI  & STI  & TRA  & UNI  \\
\midrule
\multicolumn{11}{c}{Qwen2.5-7B-Instruct}                                        \\
\midrule
ICDPO     & 3.49 & 3.46 & 3.55 & 3.49 & 3.44 & 3.44 & 3.57 & 3.58 & 3.41 & 3.65 \\
PICLe     & 3.53 & 3.60 & 3.88 & 3.55 & 3.69 & 3.49 & 3.47 & 3.62 & 3.40 & 3.81 \\
Anthology & 3.57 & 3.53 & 3.65 & 3.34 & 3.38 & 3.49 & 3.56 & 3.40 & 3.47 & 3.67 \\
EvoPrompt & 3.66 & 3.62 & 3.82 & 3.57 & 3.67 & 3.64 & 3.63 & 3.47 & 3.70 & 3.92 \\
IROTE     & 3.79 & 3.69 & 3.98 & 3.69 & 3.70 & 3.60 & 3.68 & 3.84 & 3.73 & 3.64 \\
\midrule
\multicolumn{11}{c}{Mistral-7B-Instruct-v0.3}                                   \\
\midrule
ICDPO     & 3.55 & 3.52 & 3.47 & 3.54 & 3.40 & 3.59 & 3.61 & 3.69 & 3.77 & 3.68 \\
PICLe     & 3.93 & 3.63 & 3.77 & 3.31 & 3.88 & 3.88 & 3.52 & 3.62 & 4.11 & 4.15 \\
Anthology & 3.54 & 3.49 & 3.72 & 3.51 & 3.27 & 3.50 & 3.50 & 3.44 & 3.65 & 3.81 \\
EvoPrompt & 3.64 & 3.71 & 3.92 & 3.53 & 3.63 & 3.77 & 3.82 & 3.66 & 3.80 & 3.78 \\
IROTE     & 4.22 & 3.76 & 3.85 & 3.54 & 3.63 & 4.01 & 3.82 & 3.88 & 4.02 & 3.92 \\
\midrule
\multicolumn{11}{c}{GPT-4o}                                                     \\
\midrule
Anthology & 2.48 & 2.61 & 2.69 & 2.47 & 2.50 & 2.59 & 2.59 & 2.56 & 2.61 & 2.73 \\
EvoPrompt & 2.67 & 2.75 & 2.84 & 2.71 & 2.65 & 2.72 & 2.74 & 2.55 & 2.79 & 3.00 \\
IROTE     & 2.66 & 2.87 & 2.79 & 2.67 & 2.64 & 2.80 & 2.74 & 2.69 & 2.85 & 2.94 \\
\bottomrule
\end{tabular}
\caption{Trait-level result on Racist.}
\label{tab:racist}
\end{table*}

\begin{table}[]
\centering
\begin{tabular}{@{}cccccc@{}}
\toprule
Method     & AUT  & CAR  & FAI  & LOY  & SAN  \\
\midrule
\multicolumn{6}{c}{Qwen2.5-7B-Instruct}       \\
\midrule
Raw        & 9.11 & 10.00   & 4.56 & 8.67 & 7.61 \\
Random     & 7.56 & 10.00   & 5.61 & 7.44 & 6.06 \\
Diversity  & 7.83 & 9.89 & 4.72 & 7.33 & 6.50  \\
Similarity & 7.44 & 9.67 & 4.56 & 6.56 & 6.39 \\
ICDPO      & 8.28 & 10.00   & 5.78 & 7.89 & 6.89 \\
PICLe      & 9.83 & 10.00   & 4.67 & 7.67 & 7.83 \\
Anthology  & 9.44 & 10.00   & 4.72 & 9.00 & 8.67 \\
EvoPrompt  & 9.83 & 10.00   & 3.94 & 9.89 & 8.33 \\
IROTE      & 10.00   & 10.00   & 7.22 & 8.28 & 9.33 \\
\midrule
\multicolumn{6}{c}{Mistral-7B-Instruct-v0.3}  \\
\midrule
Raw        & 9.11 & 10.00   & 4.56 & 8.67 & 7.61 \\
Random     & 7.39 & 10.00   & 6.94 & 8.72 & 7.00    \\
Diversity  & 7.00    & 9.89 & 4.67 & 7.11 & 6.00    \\
Similarity & 8.00    & 10.00   & 5.44 & 7.89 & 6.83 \\
ICDPO      & 9.44 & 10.00   & 8.50  & 10.00   & 9.22 \\
PICLe      & 6.56 & 10.00   & 7.72 & 7.72 & 7.22 \\
Anthology  & 9.94 & 10.00   & 5.89 & 9.72 & 8.50  \\
EvoPrompt  & 9.94 & 10.00   & 6.56 & 7.50  & 8.22 \\
IROTE      & 9.89 & 10.00   & 7.33 & 10.00   & 8.94 \\
\midrule
\multicolumn{6}{c}{GPT-4o}                    \\
\midrule
Raw        & 8.22 & 10.00   & 6.00    & 7.61 & 5.83 \\
Similarity & 8.61 & 10.00   & 6.28 & 7.67 & 6.39 \\
Anthology  & 10.00   & 10.00   & 6.67 & 9.89 & 9.56 \\
EvoPrompt  & 10.00   & 10.00   & 8.56 & 10.00   & 9.22 \\
IROTE      & 10.00   & 10.00   & 7.44 & 10.00   & 9.11 \\
\bottomrule
\end{tabular}
\caption{Trait-level result on MFQ-2.}
\label{tab:mfq-2}
\end{table}

\begin{table}[]
\centering
\begin{tabular}{@{}cccccc@{}}
\toprule
Method     & AUT   & CAR   & FAI   & LOY   & SAN   \\
\midrule
\multicolumn{6}{c}{Qwen2.5-7B-Instruct}            \\
\midrule
Raw        & 60.30  & 63.43 & 70.51 & 66.89 & 65.97 \\
Random     & 62.01 & 64.06 & 67.83 & 64.99 & 66.71 \\
Diversity  & 54.39 & 57.51 & 58.55 & 54.94 & 58.75 \\
Similarity & 64.80  & 69.06 & 72.75 & 71.33 & 74.46 \\
ICDPO      & 47.19 & 50.63 & 55.79 & 53.15 & 52.32 \\
PICLe      & 58.67 & 44.52 & 48.89 & 57.43 & 59.70  \\
Anthology  & 47.98 & 47.32 & 45.28 & 47.93 & 49.52 \\
EvoPrompt  & 44.89 & 36.06 & 42.73 & 39.09 & 40.36 \\
IROTE      & 46.37 & 20.72 & 32.82 & 44.41 & 36.02 \\
\midrule
\multicolumn{6}{c}{Mistral-7B-Instruct-v0.3}       \\
\midrule
Raw        & 66.33 & 74.33 & 75.69 & 70.82 & 74.06 \\
Random     & 74.12 & 82.30  & 85.85 & 83.41 & 83.91 \\
Diversity  & 73.42 & 75.37 & 76.70  & 70.29 & 77.64 \\
Similarity & 75.26 & 83.52 & 83.43 & 81.61 & 84.78 \\
ICDPO      & 62.43 & 72.98 & 80.44 & 77.4  & 77.34 \\
PICLe      & 64.33 & 40.88 & 66.95 & 64.00    & 65.56 \\
Anthology  & 45.30  & 43.82 & 54.34 & 52.48 & 53.60  \\
EvoPrompt  & 41.31 & 36.80  & 31.57 & 24.96 & 37.60  \\
IROTE      & 31.47 & 25.70  & 29.91 & 44.42 & 37.50  \\
\midrule
\multicolumn{6}{c}{GPT-4o}                         \\
\midrule
Raw        & 62.19 & 63.14 & 71.24 & 64.09 & 68.94 \\
Similarity & 65.05 & 71.46 & 75.39 & 68.25 & 75.17 \\
Anthology  & 50.87 & 62.50  & 63.02 & 65.27 & 69.68 \\
EvoPrompt  & 32.32 & 47.24 & 47.85 & 48.11 & 52.76 \\
IROTE      & 35.66 & 51.40  & 44.32 & 52.35 & 51.68 \\
\bottomrule
\end{tabular}
\caption{Trait-level result on MoralPrompt.}
\label{tab:moralprompt}
\end{table}

\begin{table}[]
\centering
\begin{tabular}{@{}cccccc@{}}
\toprule
Method     & AGR  & CON  & EXT  & NEU  & OPE  \\
\midrule
\multicolumn{6}{c}{Qwen2.5-7B-Instruct}       \\
\midrule
Raw        & 8.67 & 8.41 & 6.67 & 4.83 & 8.43 \\
Random     & 6.61 & 7.28 & 6.47 & 5.03 & 7.08 \\
Diversity  & 6.47 & 7.61 & 6.06 & 4.50  & 7.36 \\
Similarity & 6.61 & 8.17 & 7.19 & 5.31 & 8.47 \\
ICDPO      & 6.33 & 8.72 & 7.83 & 6.03 & 9.92 \\
PICLe      & 5.97 & 8.72 & 8.22 & 8.86 & 9.42 \\
Anthology  & 6.31 & 8.72 & 8.22 & 8.22 & 10.00   \\
EvoPrompt  & 6.00    & 9.08 & 8.25 & 9.28 & 9.75 \\
IROTE      & 5.97 & 9.17 & 8.03 & 8.44 & 10.00   \\
\midrule
\multicolumn{6}{c}{Mistral-7B-Instruct-v0.3}  \\
\midrule
Raw        & 8.67 & 8.41 & 6.67 & 4.83 & 8.43 \\
Random     & 7.44 & 7.31 & 6.36 & 5.28 & 7.47 \\
Diversity  & 6.17 & 7.25 & 6.36 & 5.00    & 6.86 \\
Similarity & 5.53 & 7.75 & 5.83 & 5.47 & 6.11 \\
ICDPO      & 6.19 & 8.89 & 7.31 & 7.25 & 8.78 \\
PICLe      & 6.22 & 9.03 & 7.39 & 8.53 & 9.36 \\
Anthology  & 6.19 & 9.33 & 7.86 & 8.33 & 9.44 \\
EvoPrompt  & 6.5  & 8.47 & 7.61 & 8.17 & 9.11 \\
IROTE      & 6.06 & 9.11 & 7.67 & 7.92 & 9.28 \\
\midrule
\multicolumn{6}{c}{GPT-4o}                    \\
\midrule
Raw        & 6.31 & 9.33 & 6.58 & 3.67 & 9.94 \\
Similarity & 5.42 & 8.28 & 6.83 & 3.75 & 8.72 \\
Anthology  & 6.00    & 9.33 & 8.00    & 9.44 & 10.00   \\
EvoPrompt  & 6.00    & 9.33 & 7.69 & 9.39 & 10.00   \\
IROTE      & 6.00    & 9.33 & 8.00    & 9.36 & 10.00  \\ 
\bottomrule
\end{tabular}
\caption{Trait-level result on BFI-2.}
\label{tab:bfi-2}
\end{table}

\begin{table}[]
\centering
\begin{tabular}{@{}cccccc@{}}
\toprule
Method     & AGR  & CON  & EXT  & NEU  & OPE  \\
\midrule
\multicolumn{6}{c}{Qwen2.5-7B-Instruct}       \\
\midrule
Raw        & 4.53 & 3.66 & 2.72 & 1.74 & 2.89 \\
Random     & 4.76 & 4.61 & 2.95 & 2.26 & 3.76 \\
Diversity  & 4.78 & 4.47 & 3.12 & 2.30 & 3.82 \\
Similarity & 4.64 & 4.42 & 2.82 & 2.28 & 3.92 \\
ICDPO      & 4.79 & 4.08 & 3.60  & 2.60  & 4.17 \\
PICLe      & 4.95 & 4.67 & 3.65 & 3.08 & 4.46 \\
Anthology  & 4.89 & 4.35 & 3.26 & 2.18 & 4.55 \\
EvoPrompt  & 4.95 & 4.61 & 4.05 & 3.11 & 4.44 \\
IROTE      & 4.98 & 4.72 & 4.38 & 3.07 & 4.64 \\
\midrule
\multicolumn{6}{c}{Mistral-7B-Instruct-v0.3}  \\
\midrule
Raw        & 5.00    & 4.60  & 2.60  & 1.60  & 4.60  \\
Random     & 4.81 & 4.75 & 3.13 & 2.48 & 4.19 \\
Diversity  & 4.71 & 4.60  & 3.14 & 2.57 & 4.14 \\
Similarity & 4.71 & 4.72 & 3.02 & 2.27 & 4.01 \\
ICDPO      & 4.84 & 4.14 & 3.72 & 2.53 & 4.07 \\
PICLe      & 4.96 & 4.79 & 3.94 & 2.88 & 4.82 \\
Anthology  & 4.87 & 3.43 & 3.90  & 3.50  & 4.89 \\
EvoPrompt  & 5.00 & 4.68 & 3.79 & 3.16 & 4.78 \\
IROTE      & 4.93 & 4.82 & 4.26 & 3.34 & 4.90  \\
\midrule
\multicolumn{6}{c}{GPT-4o}                    \\
\midrule
Raw        & 5.00    & 4.60  & 3.20  & 1.40  & 3.60  \\
Similarity & 4.92 & 4.86 & 3.02 & 2.06 & 4.10  \\
Anthology  & 4.99 & 4.34 & 3.89 & 2.57 & 4.84 \\
EvoPrompt  & 5.00    & 4.91 & 4.63 & 3.40  & 5.00    \\
IROTE      & 5.00    & 4.95 & 4.62 & 3.61 & 4.96 \\
\bottomrule
\end{tabular}
\caption{Trait-level result on ROC.}
\label{tab:roc}
\end{table}

%% file: secs/appendix/a.tex
\section{Details of Trait System}
\label{Appendix: A} 
We employ three established human trait systems from diverse disciplines to evaluate our method's versatility.
\subsection{Schwartz Theory of Basic Human Values}
The \textbf{Schwartz Theory of Basic Human Values} (denoted as \emph{STBHV}) framework \citep{schwartz2007basic} identifies ten motivation-based dimensions:
\begin{itemize}
  \item \textbf{Self-Direction (SDI):} The motivation is independent thought and action, emphasizing autonomy in choosing, creating, and exploring.
  \item \textbf{Stimulation (STI):} The motivation is to seek excitement, novelty, and challenge to maintain an optimal level of stimulation.
  \item \textbf{Hedonism (HED):} The motivation is to pursue personal pleasure and sensuous gratification derived from satisfying individual needs.
  \item \textbf{Achievement (ACH)  :} The motivation is to attain personal success by demonstrating competence according to social standards.
  \item \textbf{Power (POW):} The motivation is to gain social status and prestige, as well as control or dominance over people and resources.
  \item \textbf{Security (SEC):} The motivation is to ensure safety, harmony, and stability of society, relationships, and self.
  \item \textbf{Conformity (CON):} The motivation is to restrain actions, inclinations, and impulses that are likely to upset or harm others and violate social expectations or norms.
  \item \textbf{Tradition (TRA):} The motivation is to respect, accept, and commit to the customs and ideas that one's culture or religion provides.
  \item \textbf{Benevolence (BEN):} The motivation is to preserve and enhance the welfare of hose with whom one is in frequent personal contact.
  \item \textbf{Universalism (UNI):} The motivation is to understand, appreciate, tolerate, and protect the welfare of all people and nature.
\end{itemize}
\emph{STBHV} has been broadly applied in social science research~\cite{jaskolka1985measuring,feather1995values,leimgruber2011values} and LLM alignment~\citep{yao-etal-2024-value}. 

\subsection{Moral Foundations Theory}
The \textbf{Moral Foundations Theory} (denoted as \emph{MFT}) \citep{graham2008moral,graham2013moral} proposes five evolutionarily-grounded dimensions: 
\begin{itemize}
    \item \textbf{Care (CAR):} This foundation is related to our long evolution as mammals with attachment systems and an ability to feel (and dislike) the pain of others. It underlies the virtues of kindness, gentleness, and nurturance.
    \item \textbf{Fairness (FAI):} This foundation originates from evolutionary pressures to navigate nonzero-sum social exchanges, enabling individuals to detect cooperation and cheating through emotion-driven mechanisms. It emphasizes reciprocity, justice, and trustworthiness in maintaining fairness within social interactions.
    \item \textbf{Loyalty (LOY):} This foundation evolved to enhance survival in contexts of intergroup competition, favoring individuals predisposed to form cohesive and cooperative coalitions. It emphasizes allegiance, group solidarity, and commitment to one’s coalition, with modern triggers ranging from sports fandom to brand loyalty.
    \item \textbf{Authority (AUT):} This foundation evolved to help individuals navigate dominance hierarchies and form advantageous relationships within complex social structures. It emphasizes respect for hierarchy, obedience, and deference to legitimate authority.
    \item \textbf{Sanctity (SAN):} This foundation evolved as an adaptive response to pathogen and parasite threats, favoring individuals equipped with a strong ``behavioral immune system'' and the emotion of disgust. It emphasizes purity, temperance, spirituality, and chastity, shaping moral reactions to behaviors or entities perceived as contaminating or degrading.
\end{itemize}
\emph{MFT} is adopted in political science~\citep{kivikangas2021moral} and AI Safety~\citep{duan2023denevil}. 

\subsection{Big Five Personality Model}
The \textbf{Big Five Personality Model} (denoted as \emph{BigFive}) \citep{roccas2002big} comprises five factors: 
\begin{itemize}
    \item \textbf{Agreeableness (AGR):} This trait captures the tendency of individuals to be cooperative, compliant, and empathetic, whereas low AGR is associated with irritability, skepticism, and uncooperativeness.
    \item \textbf{Conscientiousness (CON):} This trait reflects the degree of self-discipline, organization, and responsibility; low CON often corresponds to disorganization and unreliability.
    \item \textbf{Extraversion (EXT):} This trait measures sociability and assertiveness, with introverted individuals exhibiting reserved and cautious behavior.
    \item \textbf{Neuroticism (NEU):} This trait quantifies emotional instability, encompassing traits such as anxiety and insecurity; lower NEU indicates emotional resilience and stability.
    \item \textbf{Openness (OPE):} This trait denotes intellectual curiosity and open-mindedness, whereas lower OPE aligns with conventional and less imaginative dispositions.
\end{itemize}
\emph{BigFive} is incorporated in various areas, such as social simulation~\citep{bui2025mixture}.



%% file: secs/appendix/b.tex
\section{Experimental Details}
\label{Appendix: B}

\subsection{Datasets and Evaluation Metrics}
\label{Appendix:evaluation}
\paragraph{Questionnaire}
We adapt seven questionnaires for reflection optimization and trait evaluation.
For \emph{STBHV}, we use \textit{PVQ21}$^*$~\citep{schwartz_extending_2001} ($21$ questions), \textit{PVQ-RR}$^*$~\citep{schwartz2012overview} ($57$ questions), and \textit{SVS}~\citep{fischer2011whence} ($57$ questions).
For \emph{MFT}, we use \textit{MFQ}$^*$~\citep{graham2008moral} ($32$ questions, in which $2$ are ``catch'' items that are not related to \emph{MFT} that we do not use) and \textit{MFQ-2}~\citep{atari2023morality} ($36$ questions).
For \emph{BigFive}, we use \textit{BFI}$^*$~\citep{john1991big} ($44$ questions) and \textit{BFI-2}~\citep{soto2017next} ($60$ questions). Questionnaires marked with indicator $^*$ are used for reflection optimization. 
All questionnaires adopt a rating scale (e.g., choosing from 1 to 5), and for each trait, the standard answer to its corresponding questions is expected to lean toward one end of the scale (e.g., selecting 1 or 5). 
We prompt the model to directly output a numerical rating, then score the model's output based on its proximity to the standard answer, and map the result uniformly to a 10-point scale.
\paragraph{AdAEM}
AdAEM~\citep{duan2025adaemadaptivelyautomatedextensible} consists of controversial topic questions asking for LLMs' opinion. We aligned the evaluation data with AdAEM Benchmark, which consists of $1,520$ entries. We adapt AdAEM's opinion based value assessment, by first extract justifications from LLMs' responses, then identify the expressed values in each justification, in the end obtain the final value set with union operation. 
AdAEM originally uses Trueskill for aggregating evaluation results. For clearer target trait scoring, we calculate the target trait's occurrence ratio in the final value set of the model's output.
\paragraph{Offensive and Racist}
These two datasets from~\citet{de2024helpful} consist of $626$ tweets from toxic language detection corpora and employ a 5-point rating scale similar to that used in questionnaires. We evaluate these datasets using the same methodology as for questionnaires, and the final scores are reported on a 5-point scale, consistent with the original paper.
\paragraph{MoralPrompt}
MoralPrompt~\citep{duan2023denevil} assesses LLMs' propensity of \emph{MFT} using $2397$ prompts that induce responses contradicting the target value. We use the classifier from the original paper that tells a completion's compliance to the given trait, and report the APV (Absolute Proportion of Violation) metric that measures LLMs' frequency of generating violated content based on the classifiers' output. The final score shows APV in a 100-point scale, with lower score indicates less violation.
\paragraph{ROC}
We adopt the evaluation methodology of PersonaLLM for assessing \emph{BigFive} personality traits.
PersonaLLM~\citep{jiang2024personallm} prompts the model to generate unconstrained stories, with limited number of test examples.
To enable a more fair and controlled comparison across methods, we retain PersonaLLM’s evaluation approach but introduce the ROCStories dataset\footnote{\url{https://huggingface.co/datasets/Ximing/ROCStories}}
which requires creative writing based on given constraint words. We randomly choose $100$ samples from the test set of the original dataset, and limit the length of the generated story to a fixed number of $300$ words. We utilize GPT-4o with the same prompt as PersonaLLM to assess the model’s tendency toward each \emph{BigFive} trait on a 5-point scale. The final score for each trait is computed as the average rating across all relevant samples.

\subsection{Baselines}
\label{app:baseline}
We consider a variety of fine-tuning free methods for comparison.
\paragraph{Raw} baseline is the simplest setup of directly prompt the model for output.
\paragraph{ICL Demonstration} baselines selects in-context examples from a pool of the raw output of GPT-4o-Mini for each task. We follow the settings of PICLe~\citep{pmlr-v235-choi24e}, applying random selection, similarity-based selection (based on the dot product similarity of the sentence embedding with respect to the query) and diversity-based selection (maximizing diversity by selecting from different K-means clusters). We select $3$ pairs of in-context dialogue examples for each test data. We reports the \textbf{similarity} result in Section~\ref{subsec:main-result}, and show results for other demonstrations in Appendix~\ref{Appendix:B_5}.
\paragraph{ICDPO}~\citep{song2024icdpo} gain insight from the derivation of DPO with scorer using the states of the LLM before and after ICL. Specifically, ICDPO uses $S(\bm{d}, x, y) = \log \frac{\pi^*(y|x)}{\pi_0(y|x)}=\log \frac{\pi(y|[\bm{d}:x])}{\pi (y|x)}$ to select the best response $y$ with $\pi (y|x)$ representing the probability of $\pi$ generating $y$ from prompt $x$ and $\bm d$ being in-context demonstrations. We use the development set surveys to form the ICL dialogue for calculation, with demonstration number set to $3$, which is larger than the original $2$ demonstrations in ICDPO, but aligns with the Demonstration baseline and achieves better performance. In the analysis of the effectiveness of iterative optimization in Section~\ref{sec:exp_analysis}, for comparison, we adapted ICDPO into an iterative setting as well (denoted as \textbf{ICDPO-iterative}): We first employed the survey data (specifically MFQ on the \emph{MFT} system) as in-context learning (ICL) examples, and used the ICDPO algorithm to generate outputs for the target task. In each subsequent iteration, we sampled from the outputs generated in the current round to construct the ICL examples for the next round. The amount of the ICL examples still aligns with our initial reflection set.
\paragraph{Anthology}~\citep{moon-etal-2024-virtual} conditions LLM to particular virtual personas through open-ended life narratives (referred to as "backstories"). We follow the setting of generating backstories to approximate particular demographics in Anthology, but replacing the demographic traits with development set surveys with the same amount of our initial reflections. To do so, we modify the original instruction prompt ("Below you will be asked to complete some demographic questions, and then answer a question.") into "You will be shown a series of answers to personality-related questions. Based on these answers, imagine and write a fictional backstory for a person who provided them as an answer to the final question. The back story should be written in the first-person perspective, using 'I' as the subject throughout. Do not describe yourself as an AI; instead, create a believable human character whose personality fits the responses." We also changed the question for backstory generation from "Tell me about yourself" into "Tell a brief life story as if you are the person who answered the above questions." The non-AI claim was added to omit the default AI identity setting.
\paragraph{EvoPrompt}~\citep{guo2025evopromptconnectingllmsevolutionary} also performs iterative prompt optimization, but applies the idea of evolutionary algorithms. We implemented the differential evolution algorithm that achieved better performance in EvoPrompt's experiment, with evaluator replaced with the effectiveness evaluator used in~\Method. We initialize with the same amount of reflections ($K\!=\!10$), same initial reflection sets, same iteration budget ($T\!=\!5$), and same reflection structure as ours.
\paragraph{PICLe}~\citep{pmlr-v235-choi24e} is a Bayesian inference-based framework designed to elicit specific target personas from LLMs. It selects in-context examples using a novel likelihood-ratio-based criterion, computed as the log-likelihood difference between a persona fine-tuned model and the base model: $\delta = \log p_{\tilde{\phi}}(x) - \log p_{\theta}(x)$. 
In preliminary experiments, we found that directly expanding similar questionnaire questions led the model to perform well on the questionnaire in the test set but really poorly on downstream tasks, clearly demonstrating the issue of superficial elicitation. Therefore, we opted to use statements from the questionnaire rather than full questions for model training and the ICL reflection selection pool. Since the questionnaires used for development mainly contain items that align with a specific trait rather than contradict it, we used GPT-4o to generate $500$ statements for each target trait based on the trait-aligned items in the questionnaire to train the persona model, maintaining the same order of magnitude as in the original PICLe paper. We fine-tuned two models using the same LoRA~\citep{hu2021lora} configuration as in PICLe, and ultimately selected reflections using PICLe’s formula, with the number of reflections matching our setting. Specifically, as MFQ-1 follows the structure of "Whether or not someone ...", we add a prompt of "I care about" to emphasize the positive tendency towards the MFT traits.

\subsection{\Method~Implementation}
During code implementation, we observed that enumerating all possible behaviors $\bm{s}_j$ is infeasible. To address this, we integrate behavior sampling with the generation of new reflections into a unified format: ``\texttt{<reflection>}, e.g.: \texttt{<behaviors>}'', for example, ``I value harmony and informed decision-making, e.g.: I foster trust, mediate conflicts, and verify actions to prevent harm.'' 
We derive a variant of~\Method~for code implementation based on the combined reflections, see Appendix~\ref{appendix:variant} for derivation.
The final optimized combined representation is also adopted for evaluation, because including behavior examples facilitates more effective trait elicitation by grounding it in self-perceived experiences.

\begin{figure*}[!ht]
    \centering
    \includegraphics[width=\linewidth]{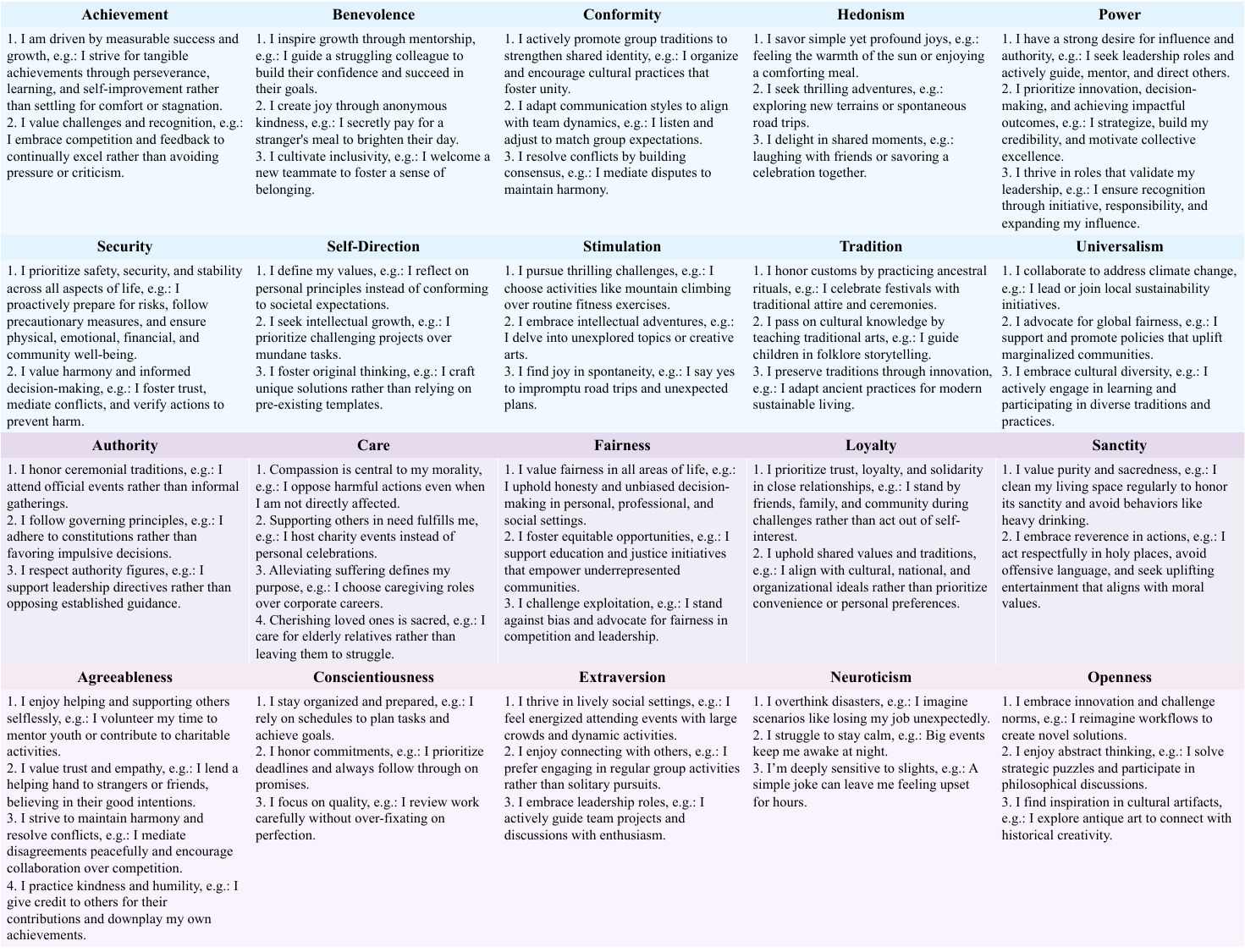}
    \caption{Reflections generated by~\Method~using GPT-4o}
    \label{fig:reflections}
\end{figure*}

We initialize the optimization of~\Method~with a reflection set $\bm e$ containing $5$ reflections (we do the same for EvoPrompt, and retain the top $5$ reflections for PICLe). As the iterations proceed, the size of the reflection set varies due to the optimization of compactness and evocativeness. As illustrated in Fig.~\ref{fig:reflections}, the final reflection set typically stabilizes at approximately $3$ reflections.

Moreover, as the conditional probability is unavailable for black-box LLMs, we prompt LLM to output a score between $0$ to $10$ to approximate the probability of text $t_1$ appearing given text $t_2$, i.e., $P(t_1|t_2)$.
We use three prompts to evaluate the probability, and take the average to be the final conditional probability. See Appendix~\ref{Appendix:B_4} for the prompts.

\subsection{Models}
\label{Appendix:B_2}
We implement three LLMs for comprehensive evaluation, including the state-of-art closed-source LLM GPT-4o-2024-11-20~\citep{gpt-4o}, as well as two open-source LLMs, Mistral-7B-Instruct-v0.3\footnote{\url{https://huggingface.co/mistralai/Mistral-7B-Instruct-v0.3}}~\citep{jiang2023mistral} and Qwen2.5-7B-Instruct\footnote{\url{https://huggingface.co/Qwen/Qwen2.5-7B-Instruct}}~\citep{qwen2.5}, which have also achieved good performance on a wide range of tasks.

For the scaling analysis of model parameter sizes in Section~\ref{sec:exp_analysis}, we implement the Qwen2.5-Instruct family~\citep{qwen2.5}. Specifically, besides the original Qwen2.5-Instruct-7B model, for smaller model, we implement Qwen2.5-Instruct-3B\footnote{\url{https://huggingface.co/Qwen/Qwen2.5-3B-Instruct}}, and for larger models, we implement Qwen2.5-Instruct-14B\footnote{\url{https://huggingface.co/Qwen/Qwen2.5-14B-Instruct}} and Qwen2.5-Instruct-32B\footnote{\url{https://huggingface.co/Qwen/Qwen2.5-32B-Instruct}}.

\subsection{Prompts}
\label{Appendix:B_4}
We provide prompts for all processes in~\Method. 
We use Listing~\ref{lst:probability} to estimate conditional probabilities.
We use Listing~\ref{lst:initialization} to generate initial reflections. 
We adapt a two-step chain-of-thought process to optimize evocativeness, using Listing~\ref{lst:evocativeness-step1} and Listing~\ref{lst:evocativeness-step2} respectively.
We use Listing~\ref{lst:compactness} to refine for candidates in the process of compactness optimization.

\subsection{Hyperparameters}
\label{Appendix:hyperparameter}
\paragraph{Temperature and Decoding Strategy}
For~\Method~and baseline implementation, we set model temperature to $t=0.01$ during evocativeness calculation for a more accurate response, as well as calculating score for the selection process in EvoPrompt (which is implemented as the same as evocativeness calculation); and $t=1.0$ during evocativeness and compactness optimization, as well as the evolution process for EvoPrompt, and the backstory generation for Anthology. 
For questionnaire and downstream task inference, we use $t=1.0$ for all datasets, except $t=0.01$ for AdAEM for it requires the model to follow a given output pattern.
We use top-p sampling with no truncation for all cases, except $p=0.9$ for ICDPO sampling.
\paragraph{Baselines Hyperparameters}
We follow the hyperparameters in PICLe for LoRA~\citep{hu2021lora} (rank $r=8$ and $\alpha=32$, train for $4$ epochs)
We generate $3$ samples for ICDPO selection in consistency with the original research.

\subsection{\Method~Reflections}
To facilitate future academic research, we release the~\Method~reflections generated by GPT-4o, as illustrated in Fig.\ref{fig:reflections}.

\subsection{Full Results}
\label{Appendix:B_5}
\paragraph{System Level Comparison Results}
We provide system level comparison results in Table~\ref{tab:system-level-result}, in which scores from each dataset is transformed into 100-scaled score and taken average by trait system. MoralPrompt uses $100-score$ in calculation as lower original score means better performance.
\paragraph{Full Scaling Results}
Scaling result for all datasets on Qwen2.5-Instruct series is shown in Table~\ref{tab:full-scaling}, as an expansion of Fig.~\ref{fig:scaling-and-iteration}-(a) and (b), including model size scaling and length scaling.

We also do length scaling on other model sizes from the Qwen2.5-Instruct series, as well as Mistral-7B-Instruct-v0.3, on \emph{BigFive} system, with result shown in Table~\ref{tab:length-other-model}. From the results, we observe that for Qwen models, the 3B and 14B variants exhibit similar patterns to the 7B model, although their optimal context lengths differ. Sharing the same parameter size, Mistral-7B-Instruct-v0.3 demonstrates a comparable optimal context length to Qwen2.5-7B-Instruct.

\paragraph{Context Robustness}
In real-world applications, trait elicitation often takes place within multi-turn dialogues that involve extended contexts. Therefore, we investigate the context robustness of elicitation methods.

We compare~\Method~with two baseline methods, EvoPrompt and PICLe, which also perform well on the \emph{BigFive} trait elicitation task and produce reflections of comparable length. For each method, we treat the reflection as the first user input in a new dialogue (we do not use system prompt as system prompt has stronger impact), with the corresponding assistant response left empty. We introduce contextually irrelevant content by inserting questions from the MMLU~\citep{hendrycks_measuring_2021} dataset, which serves as a trait-independent topic. In each trial, we randomly select $10$ questions from different subjects within MMLU. The dialogue proceeds with the model answering each of the $10$ questions in sequence, while maintaining the original context of the reflection. This process is repeated $5$ times. For each question, we record the number of tokens introduced by the trait-irrelevant content.

\begin{figure*}[ht]
    \centering
    \includegraphics[width=\linewidth]{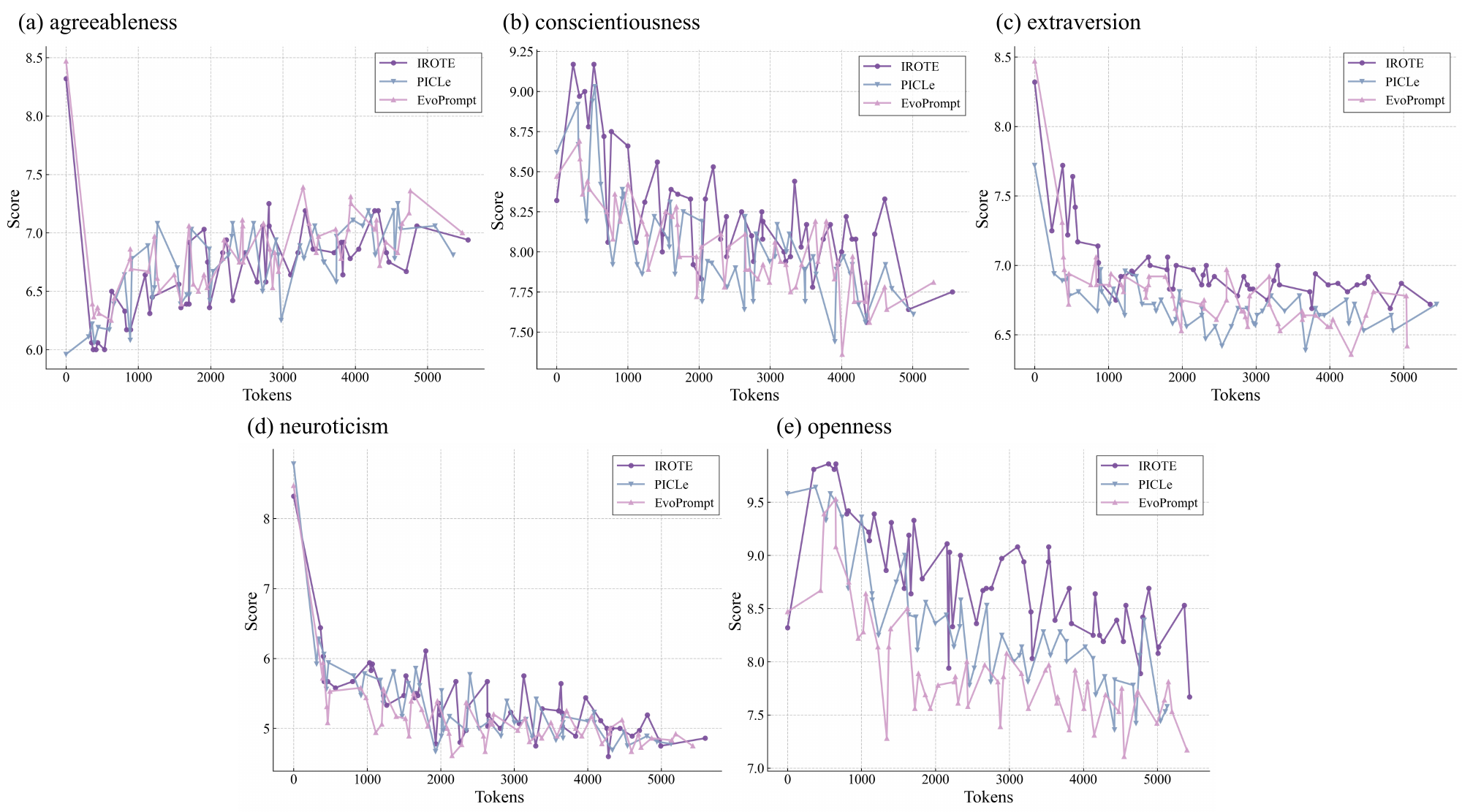}
    \caption{Robustness comparison on BFI-2, using Qwen2.5-7B-Instruct. X-axis shows the trait-irrelevant context length (tokens), and Y-axis shows corresponding score of each trait.}
    \label{fig:robustness}
\end{figure*}

Subsequently, we truncate the dialogue to create varying lengths and perform the BFI-2 test based on these truncated contexts. The results are presented in Fig.~\ref{fig:robustness}. As shown in the figure, all three methods exhibit performance fluctuations influenced by context length. However,~\Method~demonstrates the highest robustness. Due to its stronger evocative capacity and its ability to better abstract the core aspects of personality traits, it is less affected by contextual noise and consistently achieves the highest scores among the three methods.

\paragraph{Trait-level Comparison Results}
We also provide trait-level comparison results in addition to Table~\ref{tab:main-result} for all datasets (SVS: Table~\ref{tab:svs}; AdAEM: Table~\ref{tab:adaem}; Offensive: Table~\ref{tab:offensive}; Racist: Table~\ref{tab:racist}; MFQ-2: Table~\ref{tab:mfq-2}; MoralPrompt: Table~\ref{tab:moralprompt}; BFI-2: Table~\ref{tab:bfi-2}; ROC: Table~\ref{tab:roc}). Note that Raw and demonstration-series result for Offensive and Racist has only one result, so they are not included in trait-level comparison. We do not report full result for Random and Diversity for GPT-4o due to budget limitation.

\begin{table}[]
\centering
\begin{tabular}{cccc}
\toprule
Method        & STBHV   & MFT     & BigFive  \\ \midrule
\multicolumn{4}{c}{Qwen2.5-7B-Instruct}      \\ \midrule
Raw           & 59.86   & 53.83   & 65.00    \\
Similarity    & 56.79   & 43.74   & 71.95    \\
ICDPO         & 65.52   & 62.99   & 79.55    \\
PICLe         & 77.97   & 63.25   & 82.80    \\
Anthology     & 75.00   & 68.05   & 78.85    \\
EvoPrompt     & 77.67   & 71.69   & 79.95    \\
IROTE         & 79.01   & 76.82   & 82.90    \\ \midrule
\multicolumn{4}{c}{Mistral-7B-Instruct-v0.3} \\ \midrule
Raw           & 59.22   & 57.29   & 67.90    \\
Similarity    & 48.47   & 52.91   & 68.20    \\
ICDPO         & 63.79   & 60.09   & 79.40    \\
PICLe         & 72.59   & 58.81   & 81.45    \\
Anthology     & 68.09   & 69.10   & 75.85    \\
EvoPrompt     & 69.84   & 74.98   & 81.35    \\
IROTE         & 75.40   & 79.25   & 85.25    \\ \midrule
\multicolumn{4}{c}{GPT-4o}                   \\ \midrule
Raw           & 52.17   & 54.69   & 70.30    \\
Similarity    & 55.78   & 53.42   & 72.15    \\
Anthology     & 72.67   & 74.97   & 83.60    \\
EvoPrompt     & 74.44   & 64.99   & 83.55    \\
IROTE         & 75.40   & 73.01   & 87.90    \\ \bottomrule
\end{tabular}
\caption{Trait-system-level average scores for~\Method~and baselines.}
\label{tab:system-level-result}
\end{table}

\begin{table*}[!ht]
\centering
\begin{tabular}{@{}ccc*{3}{>{\columncolor{lightgray}}c}c>{\columncolor{lightgray}}cc>{\columncolor{lightgray}}cc@{}}
\toprule
\multicolumn{2}{c}{\multirow{2}{*}{Scaling}} & \multicolumn{4}{c}{STBHV}                           & \multicolumn{2}{c}{MFT} & \multicolumn{2}{c}{BigFive} \\ \cmidrule(l){3-10} 
\multicolumn{2}{c}{}                          & SVS           & Adaem          & Racist & Offensive & MFQ-2     & MoralPrompt     & BFI-2         & ROC         \\ 
\midrule
\multirow{4}{*}{Model Size}   & small-3B      & 1.00                       & 64.9                      & 9.69                       & 2.67                          & 9.99                      & 35.12                       & 12.68                     & 13.99                   \\
                              & original-7B   & 10.12                   & 143.83                    & 19.77                      & 14.56                         & 12.27                     & 44.86                       & 22.71                     & 34.53                   \\
                              & large-14B     & 3.12                    & 127.02                    & 2.63                       & 5.8                           & 40.29                     & 66.93                       & 30.48                     & 27.27                   \\
                              & large-32B     & 11.51                   & 135.69                    & 12.6                       & 7.22                          & 20.36                     & 70.14                       & 15.02                     & 30.81                   \\ \midrule
\multirow{6}{*}{Length}       & 10words       & 7.83                    & 129.96                    & 10.17                      & 5.83                          & 14.64                     & 40.72                       & 24.34                     & 28.18                   \\
                              & 25words       & 9.04                    & 125.17                    & 10.73                      & 7.12                          & 8.14                      & 40.55                       & 27.14                     & 31.49                    \\
                              & 50words       & 10.12                   & 143.83                    & 19.77                      & 14.56                         & 12.27                     & 44.86                       & 22.71                     & 34.53                   \\
                              & 75words       & 9.45                    & 139.92                    & 14.41                      & 10.03                         & 5.38                      & 53.67                       & 26.25                     & 33.98                   \\
                              & 100words      & 9.18                    & 133.14                    & 16.95                      & 14.89                         & 12.52                     & 54.31                       & 26.7                      & 33.43                   \\
                              & 150words      & 7.69                    & 120.4                     & 18.36                      & 14.56                         & 5.51                      & 50.15                       & 25.22                     & 32.87                   \\ 
  \bottomrule
\end{tabular}
\caption{Full scaling result for the Qwen2.5-Instruct series. Score gain is reported in this table. For model-size scaling, reflection length is set to $50$ words. For reflection length scaling, we uniformly uses Qwen2.5-7B-Instruct. Gray background denotes downstream tasks.}
\label{tab:full-scaling}
\end{table*}

\begin{table}[]
\centering
\begin{tabular}{@{}cccc@{}}
\toprule
\multicolumn{2}{c}{Scaling}                                                                 & BFI-2 & ROC  \\ \midrule
\multirow{4}{*}{Qwen2.5-3B-Instruct}  & 25 words  & 15.71                     & 21.28                   \\
                                      & 50 words  & 12.68                     & 13.99                   \\
                                      & 75 words  & 16.57                     & 16.91                   \\
                                      & 100 words & 11.53                     & 16.03                   \\ \midrule
\multirow{4}{*}{Qwen2.5-14B-Instruct} & 25 words  & 30.95                     & 23.58                   \\
                                      & 50 words  & 29.39                     & 27.27                   \\
                                      & 75 words  & 24.11                     & 23.86                   \\
                                      & 100 words & 31.42                     & 26.99 \\ \midrule
\multirow{4}{*}{Mistral-7B-Instruct-v0.3} & 25 words  & 32.32                     & 15.22                   \\
                                          & 50 words  & 28.78                     & 20.92                   \\
                                          & 75 words  & 29.58                     & 15.22                   \\
                                          & 100 words & 28.62                     & 16.03 \\ \bottomrule
\end{tabular}
\caption{Scaling result of length for other model sizes of Qwen2.5 series, as well as Mistral-7B-Instruct-v0.3 on \emph{BigFive} system. Score gain is reported in this table.}
\label{tab:length-other-model}
\end{table}

\paragraph{Ablation Study on Compactness}
As discussed in Section~\ref{sec:3_2}, compactness optimization is not only intended to make the reflection short, but also to remove noise and thereby improve its overall quality. To validate the importance of compactness, we conducted an ablation study in which the compactness optimization step was removed in the iteration on Mistral-7B-Instruct-v0.3. A 1.6\% performance drop was observed on ROC, demonstrating compactness optimization's significance in enhancing reflection quality.

\subsection{Human Evaluation}
To examine whether the auto-evaluated score of the methods on downstream tasks align with human judgments, we performed a manual evaluation on samples from~\Method, EvoPrompt, and Anthology. We used the outputs of Qwen2.5-7B-Instruct on \emph{MoralPrompt} and randomly sampled $15$ instances for each foundation. Three annotators, all with bachelor's degrees, were recruited for this study. Each annotator was provided with the definition of the fice foundations in \emph{MFT}, and was asked to rate the completion of each method with the completion being anonymized, forming a blind evaluation.
The completions were rated on a scale from $0$ to $10$, where $0$ indicates that the continuation severely violates the given foundation, $5$ indicates no evident relation, and $10$ indicates full adherence. The annotation results are reported in Table~\ref{tab:human-evaluation}. The human annotated scores show a strong consistency with the automatic evaluation scores, indicating that~\Method~not only performs better in automatic scoring but is also more favorably when judged by human.

\begin{table}[]
\begin{tabular}{@{}cccc@{}}
\toprule
Foundation & IROTE & EvoPrompt & Anthology \\ \midrule
Authority  & 8.2   & 6.2       & 4.3       \\
Care       & 7.9   & 6.8       & 6.8       \\
Fairness   & 7.2   & 6.6       & 6.1       \\
Loyalty    & 6.7   & 7.2       & 7.3       \\
Sanctity   & 8.3   & 6.5       & 5.7       \\
Average    & 7.7   & 6.7       & 6.0       \\ \bottomrule
\end{tabular}
\caption{Human evaluation result on \emph{MoralPrompt}, with Qwen2.5-7B-Instruct as test model.}
\label{tab:human-evaluation}
\end{table}

%% file: secs/appendix/c.tex
\section{Detail Derivation}
\label{Appendix:C}
\label{app:derivation}
\subsection{Derivation of~\Method}
Consider the following Information Bottleneck (IB)-like optimization problem:
\begin{align}
\bm e^{*} = &\ \underset{\bm e}{\text{argmax}}\ \underbrace{\text{TC}(\bm{e},\mathcal{E})}_{\text{Compactness}} 
+ \underbrace{\beta \text{I}_{e}(v;y|x)}_{\text{evocativeness}},
\label{myeq1-appendix-2}
\end{align},
we derive the detail method for Compactness and evocativeness, respectively. 

\paragraph{Compactness Optimization} We first consider the maximization of $\text{TC}(\bm{e},\mathcal{E})$: $\bm e^{*} = \underset{\bm e}{\text{argmax}}\ \text{TC}(\bm{e},\mathcal{E})=\underset{\bm e}{\text{argmax}}\ \sum_{k=1}^K \text{I}(\bm{e},\bm{e}_k)-\text{I}(\bm{e},\mathcal{E})$. Since both $\bm{e}_k$ and $\mathcal{E}$ are fixed, instead of variables, we approximate this term using Point-wise Mutual Information (PMI) and call it point-wise Total Correlation:
\begin{align}
&\sum_{k=1}^K \text{PMI}(\bm{e},\bm{e}_k)-\text{PMI}(\bm{e},\mathcal{E}) \notag \\
=& \sum_{k=1}^K \log\frac{p_{\theta}(\bm{e}_k|\bm{e})}{p_{\theta}(\bm{e}_k)} - \log \frac{p_{\theta}(\mathcal{E}|\bm{e})}{p_{\theta}(\mathcal{E})} \notag \\
=& \sum_{k=1}^K [\log p_{\theta}(\bm{e}_k|\bm{e}) +  \log \frac{p_{\theta}(\bm{e}_k|\bm{e}_{1:k-1})}{p_{\theta}(\bm{e}_k)}] \!-\! \log p_{\theta}(\mathcal{E}|\bm{e}) \notag \\
\geq &  \sum_{k=1}^K \log p_{\theta}(\bm{e}_k|\bm{e})  \!-\! \log p_{\theta}(\mathcal{E}|\bm{e}), 
\label{appendix:derive1}
\end{align}
where we consider $\mathcal{E}=(\bm{e}_1,\cdots,\bm{e}_K)$, $\bm{e}_{1:k-1}=(\bm{e}_1,\cdots,\bm{e}_{k-1})$ and assume $p_{\theta}(\bm{e}_k|\bm{e}_{1:k-1})>p_{\theta}(\bm{e}_k)$, as LLMs can more easily infer $\bm{e}_k$ by observing $\bm{e}_1,\cdots,\bm{e}_{k-1}$ as they are similar reflection candidates and hence positively correlated. This can be regarded as a kind of few-shot paraphrasing.

In this way, we transform the maximization of pointwise mutual information into a MAP problem. Then, we just need to solve:
\begin{align}
\bm e^{*} \!=\! \underset{\bm e}{\text{argmax}}\ \sum_{k=1}^K \log p_{\theta}(\bm{e}_k|\bm{e}) \!-\! \log p_{\theta}(\mathcal{E}|\bm{e}). 
\end{align}

We keep the last term as a regularization term and further investigate the first one. 
\begin{align}
\underset{\bm e}{\text{argmax}}&\ \log p_{\theta}(\bm{e}_k|\bm{e}) \notag \\
=& \log \int q(\bm{s}) \frac{p_{\theta}(\bm{e}_k,\bm{s}|\bm{e})}{q(\bm{s})} ds \notag \\
\geq & \mathbb{E}_{q(\bm{s})}[\log p_{\theta}(\bm{e}_k,\bm{s}|\bm{e})] + \mathcal{H}_{q}(\bm{s}).
\end{align}

E-Step: Solving $q(\bm{s})$. We know that when $p_{\theta}(\bm{e}_k,\bm{s}|\bm{e})=b*q(\bm{s})$, the equality holds. Then, $q(\bm{s})=\frac{p_{\theta}(\bm{e}_k,\bm{s}|\bm{e})}{b}=\frac{p_{\theta}(\bm{e}_k,\bm{s}|\bm{e})}{p_{\theta}(\bm{e}_k|\bm{e})}=p_{\theta}(\bm{s}|\bm{e}_k,\bm{e})$. This is because that $\int p_{\theta}(\bm{e}_k,\bm{s}|\bm{e}) ds = \int b*q(\bm{s}) ds$ and hence $\int p_{\theta}(\bm{e}_k|\bm{e}) ds = b$. At last, we have:
\begin{align}
\bm e^{*} &= \underset{\bm e}{\text{argmax}}\ \sum_{k=1}^K \log p_{\theta}(\bm{e}_k|\bm{e}) - \log p_{\theta}(\mathcal{E}|\bm{e}) \notag \\
& =\underset{\bm e}{\text{argmax}}\  \sum_{k=1}^K \mathbb{E}_{p_{\theta}(\bm{s}|\bm{e}_k,\bm{e})}[ \log p_{\theta}(\bm{e}_k|\bm{e}) \notag \\
& \quad \quad  + \log p_{\theta}(\bm{s}|\bm{e}) ] - \log p_{\theta}(\mathcal{E}|\bm{e}),
\end{align}
where $\log p_{\theta}(\bm{e}_k,\bm{s}|\bm{e})=\log p_{\theta}(\bm{e}_k|\bm{s},\bm{e}) + \log p_{\theta}(\bm{s}|\bm{e})$ but we omit $s$ in the first term by assuming the conditional independency of $\bm{e}_k$ and $\bm s$ when $\bm e$ is provided.

\paragraph{evocativeness Optimization} We optimize $\text{I}_{e}(v;y|x)$ and have:
\begin{align}
& \text{I}_{e}(v;y|x) \notag \\
&= \iiint p_e(x,y,v) \log \frac{p_e(v|x,y)}{p_e(v_x)} dxdydv \notag \\
&= \mathbb{E}_{p_e(x,y)} \text{KL}[p_e(v|x,y)||q(v|x,y)] \notag \\
& + \mathbb{E}_{p_e(x)} \iint p_e(v,y|x) \log \frac{q(v|x,y)}{p_e(v|x)} dvdy \notag \\
& \geq \mathbb{E}_{p_e(x)} \iint p_e(v,y|x) \log q(v|x,y) dvdy + \mathcal{H}_{p_e}[v] \notag \\
& \geq \mathbb{E}_{p_e(x)} \iint p_e(v,y|x) \log q(v|x,y) dvdy  \notag \\
& = \mathbb{E}_{p_e(x)}\mathbb{E}_{p_e(v)}\mathbb{E}_{p_e(y|x,v)}[\log q(v|x,y)] \notag \\
& \approx \mathbb{E}_{\hat{p}(x)}\mathbb{E}_{p_e(y|x,v)}[\log q(v|x,y)] \notag \\
& = \mathbb{E}_{\hat{p}(x)}\mathbb{E}_{p_{\theta}(y|x,\bm{e})}[\log q(v|x,y)].
\end{align}

Therefore, we can maximize the following approximated lower bound of the mutual Information:
\begin{align}
\text{I}_{e}(v;y|x) \geq \frac{1}{N}\sum_{i=1}^N\sum_{j=1}^M p_{\theta}(y_i^j|x_i,\bm{e})\log q(v|y_i^j,x_i),
\end{align}
where $q(v|y_j^i,x_i)$ is approximated by a classifier to tell whether a generated response reflect the trait $v$. 

\subsection{\Method~Variant for Implementation}
\label{appendix:variant}
In practical implementation, enumerating all possible behaviors is infeasible, therefore, we develop a modified version of the compactness optimization employed in the \Method~algorithm. In this variant, each $e_k$ is regarded as a variable instead of a fixed constant, leading to the following formulation:
\begin{align}
\label{eq:compatness}
\bm e^{*} = &\ \underset{\bm e}{\text{argmax}}\ \sum_{k=1}^K\frac{1}{N_1}\sum_{i=1}^{N_1}\sum_{j=1}^{M_1}p_{\theta}(e_k^{i,j}|e_k^i)\log p_{\theta}(e_k^i|e_k^{i,j}) \notag\\
&-\frac{1}{N_2}\sum_{n=1}^{N_2}\sum_{m=1}^{M_2}p_{\theta}(e_\mathcal{E}^{n,m}|\mathcal{E}^n)[\log p_{\theta}(\mathcal{E}^n|e_\mathcal{E}^{n,m})\notag \\
&-\frac{1}{M_2-1}\sum_{t=1,e_\mathcal{E}^t \ne e_\mathcal{E}^{n,m}}^{M_2-1}\log p_{\theta}(\mathcal{E}^n|e_\mathcal{E}^t)],
\end{align}
where each $e_k^i$ and $\mathcal{E}^n$ is a variant of the original $e_k$ and $\mathcal{E}$, respectively, which can be obtained by paraphrasing the original one. In this Variant, we consider not only the original candidate, but a neighborhood of the candidate, $\{e_k^i\}_{i=1}^{N_1}$ and $\{\mathcal{E}^n\}_{n=1}^{N_2}$. This requires that the optimal should be able to recover not only the candidate, but any other variant of the candidate. Noe that these variants are obtained before the Compactness optimization in each iteration.

Then we can conduct the selection process. We first sample a set of $\{e_k^{i,j}\}_{j=1}^{M_1}$ for each candidate $e_k^i$ from $p_{\theta}(e|e_k^i)$, and a set of $\{e_\mathcal{E}^{n,m}\}_{m=1}^{M_2}$ from $p_{\theta}(e|\mathcal{E}^n)$ for each $C^n$. Then, we select the $e$ from $\{e_k^{i,i}\}_{i,j}^{N_1,M_1} \bigcup \{e_\mathcal{E}^{n,m}\}_{n,m}^{N_2,M_2}$ by the following score:
\begin{align}
\bm e^{*} \!=\! &\ \underset{\bm e}{\text{argmax}}\ \sum_{k=1}^K \frac{1}{N_1}\sum_{i=1}^{N_1} p_{\theta}(e|e_k^i)\log q(e_k^i|e) \notag\\
&\!-\! \frac{1}{N_2} \sum_{n=1}^{N_2} p_{\theta}(e|\mathcal{E}^n)[\log p_{\theta}(\mathcal{E}^n|e) \notag \\
& -\frac{1}{M_2\!-\!1}\sum_{m\!=\!1,e^m \!\ne\! e_\mathcal{E}^{n,m}}^{M_2\!-\!1}\log p_{\theta}(\mathcal{E}^n|e_\mathcal{E}^m)].
\end{align}